\documentclass[10pt,twocolumn,letterpaper]{article}

\usepackage[pagenumbers]{cvpr}      

\newcommand{\red}[1]{{\color{red}#1}}

\usepackage{multirow}
\usepackage{colortbl}
\definecolor{oursrow}{RGB}{228,230,250}
\newcommand{\best}[1]{\textbf{#1}}
\newcommand{\second}[1]{\underline{#1}}

\definecolor{cvprblue}{rgb}{0.21,0.49,0.74}
\usepackage[pagebackref,breaklinks,colorlinks,allcolors=cvprblue]{hyperref}
\usepackage{booktabs,graphicx,pifont}
\usepackage{adjustbox}
\usepackage[ruled,linesnumbered]{algorithm2e}
\newcommand{\maskfree}{\ding{51}}
\newcommand{\maskguided}{\ding{55}}

\usepackage{array}
\usepackage{tabularx}
\usepackage{marvosym}
\def\confName{CVPR}
\def\confYear{2027}
\usepackage{amsmath}
\usepackage{amssymb}
\usepackage{mathtools}
\usepackage{amsthm}
\title{Taming Video Generative Models for One-Step Mask-Free Video Overlay Removal}

\author{%
  \textbf{Zijie Lou}\textsuperscript{1*},
  \textbf{Xiangwei Feng}\textsuperscript{1*},
  \textbf{Jifei Che}\textsuperscript{1},
  \textbf{Xiaochao Qu}\textsuperscript{1},
  \textbf{Luoqi Liu}\textsuperscript{1},
  \textbf{Ting Liu}\textsuperscript{1\Letter}\\
  {\normalsize
    \textsuperscript{1}\,MT Lab, Meitu Inc.
    \quad \textsuperscript{*}\,Equal contribution
    \quad \textsuperscript{\Letter}\,Corresponding author}
}

\begin{document}
\maketitle
\begin{abstract}
Video overlay removal aims to recover clean footage obscured by overlays such as subtitles, watermarks, and logos. Existing methods often require multiple inference steps and input masks. We present SmartEraser, which adapts a video generative model for one-step mask-free overlay removal. Because overlays typically leave much of the original scene visible, SmartEraser starts restoration directly from the overlaid video. We formulate restoration as residual regression in the model’s representation space using Overlay-to-Clean Residual Matching (OCRM). The model learns restoration corrections from stochastic interpolants between the representations of paired overlaid and clean videos. At inference, the model restores the overlaid video in a single step. Synthetic training pairs provide ground-truth masks, but accurate masks are difficult to obtain at test time. During training, we apply the Privileged Mask Curriculum (PMC) learning strategy, which initially uses these masks as spatial guidance and gradually reduces their availability. As mask guidance is withdrawn, the model learns to restore videos without input masks. We also introduce SmartEraser-Bench, which contains 1,000 synthetic videos simulating real-world overlay scenarios across ten categories and 100 real-world videos collected online. Extensive experiments demonstrate that our approach outperforms existing methods in both visual quality and temporal consistency.

\end{abstract}    
\section{Introduction}
Video overlay removal is the task of eliminating composited graphics from video frames and reconstructing the affected scene content without disturbing the surrounding footage. Successful removal therefore requires both identifying the overlay-affected regions and recovering their content with spatial fidelity and temporal coherence.

Existing approaches commonly formulate this task as video inpainting. ProPainter \cite{zhou2023propainter} reconstructs specified regions by combining temporal propagation with mask-guided content completion. Generative methods such as VideoPainter \cite{bian2025videopainter}, DiffuEraser \cite{li2025diffueraser}, and MiniMax-Remover \cite{zi2026minimax} leverage learned video priors to infer the affected content. These methods require input masks to indicate where restoration is needed, making removal quality sensitive to mask coverage. When an overlay extends beyond the specified regions, unmarked fragments may remain in the output. Moreover, existing methods \cite{bian2025videopainter, li2025diffueraser, zi2026minimax, miao2026deoverlay} typically require multiple inference steps, increasing the computational cost of restoration.

\begin{figure}[!t]
\centering
\includegraphics[width=\linewidth]{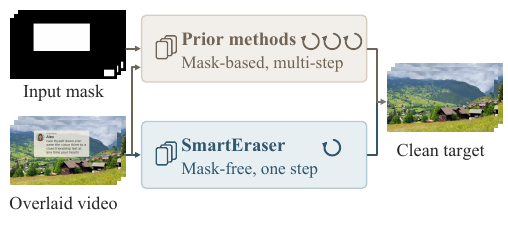}
\caption{Comparison of previous and our video overlay removal framework.}
\label{diff}
\end{figure}

Recent subtitle removal methods have sought to reduce these requirements. CLEAR \cite{he2026clear} operates without externally supplied masks, while SEDiT \cite{hui2026sedit} supports both mask-free inference and single-step sampling. However, complex overlay styles and rendering conditions continue to limit their removal quality. CLEAR can struggle with artistic titles, and SEDiT's base model can leave visible subtitle fragments in challenging real videos. These limitations highlight the need to combine low inference cost with reliable restoration across a broader range of overlays.

Existing datasets typically use paired overlaid and clean videos to support quantitative evaluation, with controlled variations in overlay appearance and placement \cite{he2026clear,hui2026sedit,miao2026deoverlay}. However, their coverage of complex compositions involving multiple overlay elements and editing effects remains limited. For example, live-chat panels can combine scrolling comments, emoji, and translucent backgrounds, while shot transitions can blur these elements together with the underlying footage. Such conditions call for broader evaluation of interactions among overlay content, rendering effects, and scene dynamics.

To address these challenges, we present SmartEraser, a framework for direct restoration of overlaid videos without inference-time masks. Our design uses the encoded input as the starting state of restoration, taking advantage of the scene information that remains visible despite the overlays. We adapt a pretrained video generative model through Overlay-to-Clean Residual Matching. During training, stochastic interpolants connect the latent representations of paired overlaid and clean clips, and the model learns to estimate restoration corrections from these intermediate states. At test time, the model predicts a correction from the encoded overlaid input and applies it once, replacing iterative sampling with a single restoration update.

To provide spatial guidance without making masks a test-time requirement, we introduce the privileged mask curriculum learning strategy. Ground-truth masks from synthetic pairs initially indicate the regions that require modification. We then progressively reduce the probability of mask conditioning until masks are no longer supplied. Paired clean clips continue to supervise restoration throughout this transition. The curriculum thus provides explicit localization early in training and concludes under the same mask-free input condition used at inference.

We further construct SmartEraser-Bench to evaluate restoration under diverse overlay conditions. The benchmark comprises 1,000 synthetic videos and 100 real-world videos collected online. The synthetic set covers 10 common overlay scenarios, such as glowing subtitles, scrolling live chat, dynamic watermarks, emoji, stickers, and transition blur. Each synthetic video is paired with a clean reference for controlled quantitative comparison. The real-world videos are accompanied by manually annotated overlay masks for evaluation and provide a direct test of generalization beyond synthetic overlay conditions. We will make SmartEraser-Bench publicly available to support reproducible evaluation of video overlay removal methods.

Our contributions are summarized as follows:
\begin{itemize}
\item{We propose SmartEraser, a one-step mask-free video overlay removal framework through overlay-to-clean residual matching and privileged mask curriculum learning strategy.}

\item{We construct SmartEraser-Bench, containing 1,000 synthetic videos across 10 common overlay scenarios and 100 real-world videos collected online. The benchmark combines reference-based evaluation on synthetic videos with generalization assessment on real-world videos, for which we provide manually annotated overlay masks.}

\item{Extensive experiments demonstrate that SmartEraser achieves state-of-the-art overlay removal performance on both synthetic and real-world videos.}
\end{itemize}
\begin{table}[t]
  \centering
  \caption{\textbf{Comparison of video overlay removal methods.}
  Source initialization means that sampling starts from the latent of
  the overlaid input video.}
  \label{tab:inference_design}
  \resizebox{\linewidth}{!}{%
    \begin{tabular}{lccc}
      \toprule
      Method
      & \shortstack{Source\\initialization}
      & \shortstack{One-step\\Inference}
      & \shortstack{Mask-free} \\
      \midrule
      SEDiT~\cite{hui2026sedit}     & \ding{55} & \ding{51} & \ding{51} \\
      CLEAR~\cite{he2026clear}      & \ding{55} & \ding{55} & \ding{51} \\
      DeOverlay~\cite{miao2026deoverlay}
                                   & \ding{51} & \ding{55} & \ding{55} \\
      \midrule
      \rowcolor{oursrow}
      SmartEraser (Ours)  & \ding{51} & \ding{51} & \ding{51} \\
      \bottomrule
    \end{tabular}%
  }
\end{table}
\section{Related Work}
\subsection{Video Generative Models}
Recent video generative models combine spatiotemporal compression with transformer-based sequence modeling. LTX-Video \cite{hacohen2024ltx} emphasizes generation efficiency through highly compressed representations and coordinated autoencoder and denoiser design. HunyuanVideo \cite{kong2024hunyuanvideo} and Wan-Video \cite{wan2025} further explore large-scale training through advances in data processing, model architecture, and training infrastructure. These models provide reusable priors for scene appearance, structure, and motion. However, they typically require multiple inference steps, which increases the computational cost of generation. Adapting these capabilities to restoration requires preserving observed content while reducing inference cost. Existing video editing models \cite{bian2025videopainter,hui2026sedit} introduce source information through dedicated context encoders or reference tokens. Nevertheless, reliable one-step video editing remains challenging, and existing acceleration strategies often rely on distillation.

\subsection{Video Object and Overlay Removal}
Video object removal methods exploit temporal observations and learned priors to reconstruct occluded scene content. DiffuEraser~\cite{li2025diffueraser} initializes its diffusion process by applying DDIM inversion to a preliminary ProPainter reconstruction. MiniMax-Remover~\cite{zi2026minimax} uses minimax distillation to reduce the cost of sampling. These methods use masks to localize the region to be edited at inference. For video overlay removal, DeOverlay~\cite{miao2026deoverlay} initializes sampling from the source latent and applies repeated flow updates guided by the edit mask and fixed source features. CLEAR~\cite{he2026clear} learns subtitle-aware priors and adaptive reconstruction weights from paired videos and uses them for mask-free, multi-step denoising. SEDiT~\cite{hui2026sedit} performs one-step, mask-free denoising from noise, conditioning the diffusion transformer on reference-video tokens and assigning greater loss weight to subtitle regions during training. As summarized in Table~\ref{tab:inference_design}, SmartEraser learns a direct bridge from overlaid to clean video latents, combining source initialization with one-step, mask-free inference.

\begin{figure*}[!t]
\centering
\includegraphics[width=\textwidth]{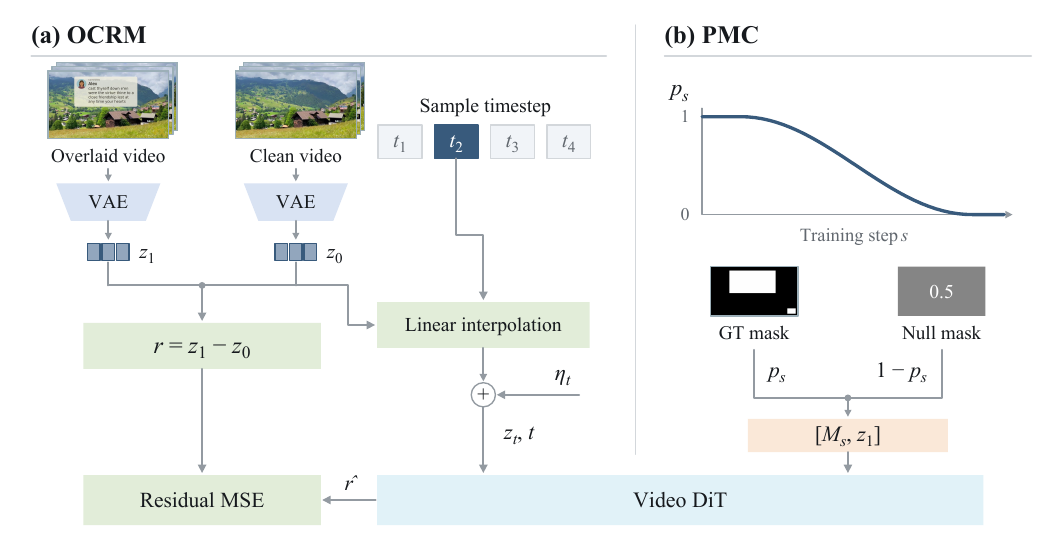}
\caption{The framework of SmartEraser. (a) OCRM draws one discrete timestep \(t\) per training example, illustrated by the highlighted cell. It linearly interpolates the overlaid latent \(z_1\) and clean latent \(z_0\) at \(t\), then adds a timestep-scaled Gaussian perturbation \(\eta_t\) to obtain \(z_t\). The residual target \(r=z_1-z_0\) is fixed for each video pair. (b) At optimizer step \(s\), \(p_s\) denotes the probability of retaining the ground-truth mask. The mask condition \(M_s\) contains the ground-truth mask with probability \(p_s\) and is filled with \(0.5\) otherwise. PMC reduces \(p_s\) from one to zero through a cosine schedule. The Video DiT predicts the residual and is optimized using mean squared error. The Video DiT also receives a T5 text embedding. At inference, we cache the embedding of a fixed prompt (\textit{Remove the text and logo in the video}).}
\label{fig:method}
\end{figure*}
\section{Method}
As shown in Fig.~\ref{fig:method}, SmartEraser adapts a video generative model for one-step, mask-free video overlay removal by Overlay-to-Clean Residual Matching (OCRM) and Privileged Mask Curriculum (PMC) learning strategy.

\subsection{Overlay-to-Clean Residual Matching}
\label{sec:ocrm}

Overlays typically leave much of the underlying scene visible. We therefore use the observed video as the starting point for restoration. Let $V_1$ and $V_0$ denote a paired overlaid and clean video, respectively. A frozen VAE encoder $\mathcal{E}$ maps them to latents $z_1=\mathcal{E}(V_1)$ and $z_0=\mathcal{E}(V_0)$. OCRM trains a video diffusion transformer $v_\theta$ to predict the residual $r=z_1-z_0$. Subtracting this residual from the source latent recovers the clean latent.

\paragraph{Training states and objective.}
As shown in Fig.~\ref{fig:method}(a), we draw one timestep $t$ uniformly from $\mathcal{T}=\{250,500,750,1000\}$ for each training example. With $\sigma_t=t/1000$, we construct the training state as
\begin{equation}
\label{eq:ocrm_state}
\begin{aligned}
z_t &= (1-\sigma_t)z_0+\sigma_t z_1+\eta_t,\\
\eta_t &= \beta\sqrt{\sigma_t(1-\sigma_t)}\,\epsilon,
\qquad \epsilon\sim\mathcal{N}(0,I),
\end{aligned}
\end{equation}
where $\beta$ controls the perturbation magnitude. The unperturbed interpolation is $z_0+\sigma_t r$, whose slope with respect to $\sigma_t$ is the constant residual $r$. The perturbation exposes the model to states near this line, while the regression target remains fixed for each video pair. At $t=1000$, the perturbation vanishes and the training state is exactly $z_1$.

The three interior timesteps provide supervision at different mixtures of the clean and overlaid latents. The fourth allocates one quarter of the training draws to the exact starting latent used at inference. This gives the model direct supervision at the source endpoint while retaining coverage of interior states. At optimizer step $s$, the prediction and training objective are
\begin{equation}
\label{eq:ocrm_loss}
\begin{aligned}
\widehat r &= v_\theta\bigl(z_t,t,[M_s,z_1],c\bigr),\\
\mathcal{L}_s(\theta)
&= \mathbb{E}\bigl[\operatorname{MSE}(\widehat r,r)\bigr],
\end{aligned}
\end{equation}
where $[M_s,z_1]$ denotes channel-wise concatenation and $c=\operatorname{T5}(P)$ is the embedding of the restoration prompt $P$. Section~\ref{sec:pmc} defines the mask condition $M_s$.

\paragraph{One-step restoration.}
At inference, we start from $z_1$ and use the spatially uniform condition $M_{\varnothing}$ defined below. A single model evaluation yields
\begin{equation}
\label{eq:ocrm_inference}
\begin{aligned}
\widehat r &= v_\theta\bigl(z_1,1000,[M_{\varnothing},z_1],c\bigr),\\
\widehat z_0 &= z_1-\widehat r,\qquad
\widehat V_0=\mathcal{D}(\widehat z_0),
\end{aligned}
\end{equation}
where $\mathcal{D}$ is the VAE decoder. Since $\widehat z_0-z_0=r-\widehat r$, the residual error at the source endpoint equals the one-step reconstruction error in latent space. The 4-point grid ensures that this endpoint is sampled during training. The direct residual update in Eq.~\eqref{eq:ocrm_inference} enables one-step inference. Algorithm~\ref{alg:smarteraser} summarizes the training and inference procedures.

\subsection{Privileged Mask Curriculum}
\label{sec:pmc}

\begin{figure}[!t]
  \centering
  \includegraphics[width=\columnwidth]{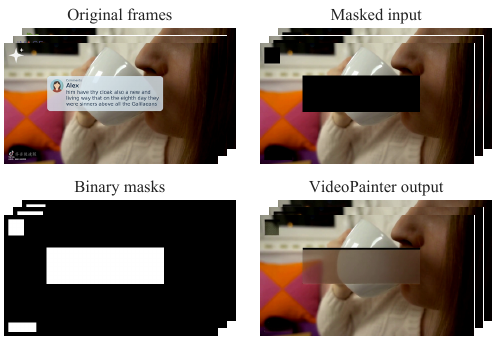}
  \caption{\textbf{VideoPainter results for video overlay removal.} VideoPainter replaces the pixels marked by the binary mask with black in the original video and then inpaints the masked regions.}
  \label{fig:videopainter_comment_card}
\end{figure}

VideoPainter, DiffuEraser, and MiniMax-Remover use binary masks to construct masked video conditions for inpainting~\cite{bian2025videopainter,li2025diffueraser,zi2026minimax}. Hard masking removes all pixel values inside the designated region from that condition. For a partially transparent overlay, these pixels can still contain scene information. As shown in Fig.~\ref{fig:videopainter_comment_card}, parts of the cup remain visible through the translucent comment card. Hard masking removes these details from the masked input. Across the three frames, VideoPainter leaves artifacts over the cup.

\begin{figure*}[!t]
\centering
\includegraphics[width=\textwidth]{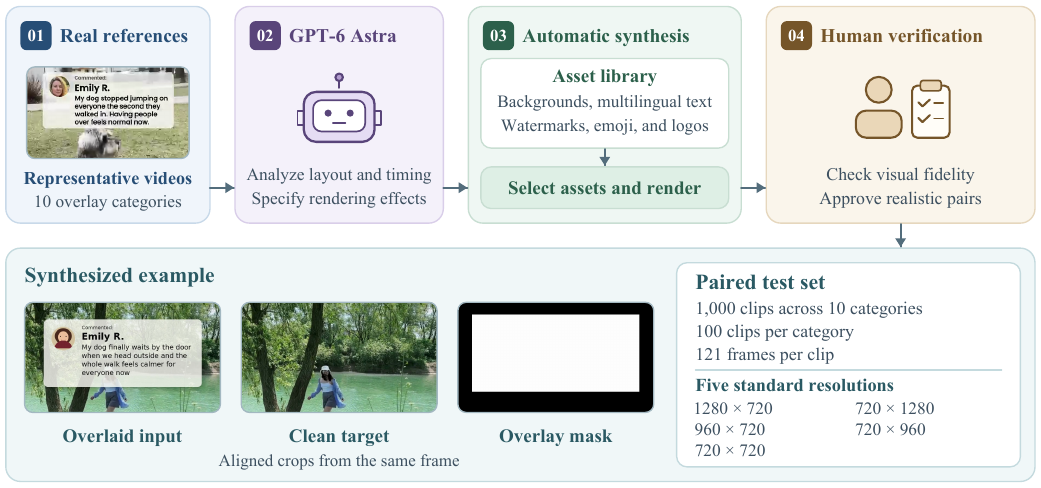}
\caption{\textbf{Synthetic video construction pipeline for SmartEraser-Bench.} Using comment cards as an example, GPT-6 Astra describes their appearance, placement, and timing to guide rendering.}
\label{testing_data_pipeline}
\end{figure*}

\begin{table*}[!t]
  \centering
  \captionsetup{width=\textwidth}
  \caption{\textbf{Comparison of video overlay removal benchmarks.} SmartEraser-Bench provides paired clean reference videos only for the synthetic split.}
  \label{tab:benchmark_comparison}
  \resizebox{\textwidth}{!}{%
    \begin{tabular}{lcccccccccc}
      \toprule
      & & \multicolumn{3}{c}{Ground Truth}
        & \multicolumn{4}{c}{Overlay Diversity} & & \\
      \cmidrule(lr){3-5}
      \cmidrule(lr){6-9}
      Benchmark & Videos
        & Clean & Mask & \shortstack{Clean First\\Frame}
        & Types & Multilingual & Spatial & Temporal
        & \shortstack{Includes\\Real-world Videos}
        & \shortstack{Train--Test\\Distribution Shift} \\
      \midrule
      CLEAR test set \cite{he2026clear} & 400
        & \ding{51} & \ding{55} & \ding{55}
        & 1 & \ding{55} & \ding{51} & \ding{55}
        & \ding{55} & \ding{55} \\
      VSR-Bench-400 \cite{hui2026sedit} & 400
        & \ding{51} & \ding{51} & \ding{55}
        & 1 & \ding{51} & \ding{51} & \ding{55}
        & \ding{55} & \ding{55} \\
      DVTE-Bench \cite{miao2026deoverlay} & 1,631
        & \ding{51} & \ding{51} & \ding{55}
        & 8 & \ding{51} & \ding{51} & \ding{51}
        & \ding{55} & \ding{55} \\
      \midrule
      \rowcolor{oursrow}
      \textbf{SmartEraser-Bench (Ours)} & \textbf{1,100}
        & \ding{51} & \ding{51} & \ding{51}
        & \textbf{10+} & \ding{51} & \ding{51} & \ding{51}
        & \ding{51} & \ding{51} \\
      \bottomrule
    \end{tabular}%
  }
\end{table*}

\paragraph{Spatial conditioning.}
Throughout this paper, an overlay mask $M$ is a binary mask derived from per-frame bounding boxes. In the training set and in both the synthetic and real-world splits of SmartEraser-Bench, a pixel is set to 1 if it lies inside any overlay bounding box and to 0 otherwise. We invert and pack $M$ onto the latent grid to obtain $\widetilde M$, where zero and one encode the masked and known states, respectively. For mask-free conditioning, we define $M_{\varnothing}$ as a tensor of the same shape with every entry set to $0.5$. This midpoint avoids assigning either binary state to the entire video. Its spatial uniformity provides no localization information, so the model must infer where to apply the correction from the video features. The full source latent $z_1$ is available under both conditions.

\begin{algorithm}[!t]
\small
\caption{SmartEraser Training and Inference}
\label{alg:smarteraser}
\KwIn{Training tuples $(V_1,V_0,M,P)$, test video $V_1^*$,
fixed removal prompt $P_*$, frozen VAE $(\mathcal E,\mathcal D)$
and T5 encoder, Video DiT $v_\theta$, schedule $\sigma_t$,
noise scale $\beta$, and step counts
$S,S_{\mathrm{warm}},S_{\mathrm{trans}}$}
\KwOut{Trained Video DiT and restored video $\widehat V_0^*$}

\textbf{Training}\;
$\mathcal T \gets \{250,500,750,1000\}$\;
\For{optimizer step $s=0,\ldots,S-1$}{
    Sample a minibatch of training tuples\;
    $z_1\gets\mathcal E(V_1)$,
    $z_0\gets\mathcal E(V_0)$,
    $c\gets\mathrm{T5}(P)$\;
    For each video, draw $t\sim\operatorname{Unif}(\mathcal T)$
    and $\epsilon\sim\mathcal N(0,I)$\;
    $\eta_t\gets\beta\sqrt{\sigma_t(1-\sigma_t)}\,\epsilon$\;
    $z_t\gets\sigma_t z_1+(1-\sigma_t)z_0+\eta_t$,
    $r\gets z_1-z_0$\;

    $\tau_s\gets\operatorname{clip}_{[0,1]}
    \bigl((s-S_{\mathrm{warm}})/(S_{\mathrm{trans}}-1)\bigr)$\;
    $p_s\gets\tfrac12(1+\cos(\pi\tau_s))$\;
    Pack $1-M$ into four channels and resize to the latent grid
    of $z_1$ to obtain $\widetilde M$\;
    Fill a tensor shaped like $\widetilde M$ with $0.5$
    to obtain $M_{\varnothing}$\;
    For each video, sample
    $M_s\gets
    \begin{cases}
        \widetilde M, & \text{with probability }p_s,\\
        M_{\varnothing}, & \text{with probability }1-p_s
    \end{cases}$\;
    $\widehat r\gets v_\theta(z_t,t,[M_s,z_1],c)$\;
    Update $\theta$ using $\operatorname{MSE}(\widehat r,r)$\;
}

\BlankLine
\textbf{Inference}\;
$z_1^*\gets\mathcal E(V_1^*)$, $c_*\gets\mathrm{T5}(P_*)$\;
Create $M_*$ with four channels on the latent grid of $z_1^*$,
all filled with $0.5$\;
$t_*\gets1000$\;
$\widehat r_*\gets v_\theta(z_1^*,t_*,[M_*,z_1^*],c_*)$\;
$\widehat z_0^*\gets z_1^*-\widehat r_*$\;
\Return{$\widehat V_0^*=\mathcal D(\widehat z_0^*)$}\;
\end{algorithm}

\paragraph{Curriculum schedule.}
We treat the ground-truth mask as privileged information available during training. Early in training, it indicates where restoration is needed. We then reduce its availability so that the model gradually learns to localize and remove overlays from the input video. As shown in Fig.~\ref{fig:method}(b), at optimizer step $s$ we sample the mask condition independently for each video:
\begin{equation}
\label{eq:pmc_condition}
M_s =
\begin{cases}
\widetilde M, & \text{with probability } p_s,\\
M_{\varnothing}, & \text{with probability } 1-p_s.
\end{cases}
\end{equation}
We retain all masks during warm-up, anneal $p_s$ with a cosine schedule, and train with $p_s=0$ after the transition. With $S_{\mathrm{warm}}$ warm-up steps and $S_{\mathrm{trans}}$ transition steps, the schedule is
\begin{equation}
\label{eq:pmc_schedule}
\begin{aligned}
\tau_s &=
\operatorname{clip}_{[0,1]}
\left(\frac{s-S_{\mathrm{warm}}}{S_{\mathrm{trans}}-1}\right),\\
p_s &= \tfrac{1}{2}\bigl(1+\cos(\pi\tau_s)\bigr).
\end{aligned}
\end{equation}

The schedule defines a continuation between two conditioning regimes. In expectation, the objective at step $s$ is
\begin{equation}
\label{eq:pmc_objective}
\mathcal{L}_s(\theta)
=
p_s\,\mathcal{L}_{\mathrm{mask}}(\theta)
+
(1-p_s)\,\mathcal{L}_{\mathrm{free}}(\theta),
\end{equation}

where $\mathcal{L}_{\mathrm{mask}}$ and $\mathcal{L}_{\mathrm{free}}$ apply the same OCRM loss with $\widetilde M$ and $M_{\varnothing}$, respectively. The curriculum changes the available spatial information while keeping the paired clean supervision and residual target fixed. The final phase uses the same mask-free condition as inference.

\section{Experiments}
\subsection{Experimental Setup}
\noindent
\textbf{Training Data.}
We construct a training set of 50,000 samples.  We generate clean backgrounds from single clips, sequences joined by hard cuts, or split screen compositions. We sample text from a multilingual corpus and vary its typography, placement, and timing. The renderer also composites graphical elements such as logos, watermarks, and emoji. For each sample, we export the overlaid video, its clean counterpart, and a mask video derived from the overlay bounding boxes in each frame. The three outputs share the same spatial dimensions and frame indices. See Appendix~\ref{training_data} for details on how we construct the training set.

\noindent
\textbf{Testing Data.}
The existing test sets listed in Table~\ref{tab:benchmark_comparison} are not publicly available. We therefore construct SmartEraser-Bench, a benchmark of 1,000 synthetic videos and 100 real-world videos. To evaluate generalization beyond the training distribution, we use a separate construction pipeline guided by representative overlay scenarios. As shown in Fig.~\ref{testing_data_pipeline}, we use GPT-6 Astra \cite{openai2026gpt6astra} to characterize the appearance, layout, and temporal behavior of representative overlays. These descriptions guide asset selection and rendering from a library of background clips, multilingual text, and graphics. Human reviewers verify that the rendered videos reproduce the intended scenarios.

The synthetic split contains 100 videos in each of 10 categories: standard and rotated subtitles, subtitles with diverse emojis and stickers, subtitles with glow effects, full-screen text overlays, comment card overlays, overlays with transition blur, live chat overlays, dynamic platform watermarks, translucent subtitles, and curved and animated subtitles. Each category is defined by its dominant overlay pattern, and clips may also contain overlays from other categories. Each synthetic sample has 121 frames and includes an overlaid input, an aligned clean target, and a per-frame overlay mask. The real-world split consists of videos collected from the Internet with naturally occurring overlays that we manually annotate in every frame. For methods that require an overlay-free first frame, such as VideoPainter~\cite{bian2025videopainter}, we use the first frame of the aligned clean target for synthetic videos and use GPT-Image \cite{openai_gpt_image} to remove any overlays present in the first frame of real-world videos. See Appendix~\ref{testing_data} for a detailed description of the benchmark.

\noindent
\textbf{Metrics.}
On the synthetic split, we evaluate overlay removal with Masked PSNR and Masked LPIPS~\cite{zhang2018unreasonable}, and background preservation with Unmasked PSNR and Unmasked LPIPS, using the aligned clean videos as references. We assess temporal behavior over the full video using Temporal Change Error (TC)~\cite{huang2026mfvsr}, Temporal Warping Error (TWE)~\cite{lai2018learning}, and mean optical flow magnitude (Flow Mean). We also report Video Fr\'echet Inception Distance (VFID)~\cite{wang2018video} between the restored and clean video sets. On the real-world split, we measure removal success and visual quality using a fixed visual review rubric, drawing on the evaluation criteria of MiniMax-Remover~\cite{zi2026minimax}. We assess background preservation with Unmasked PSNR and Unmasked LPIPS against the original input, and report CLIP-based temporal consistency (CLIP TC)~\cite{radford2021learning}, Output-flow TWE, and Flow Mean over the full video. See Appendix~\ref{app:metrics} for metric definitions and the visual review protocol.

\noindent
\textbf{Compared Methods.}
We evaluate SmartEraser against four baselines using their official code: DiffuEraser~\cite{li2025diffueraser}, VideoPainter~\cite{bian2025videopainter}, MiniMax-Remover~\cite{zi2026minimax}, and CLEAR~\cite{he2026clear}. We also fine-tune the three baselines with publicly available training code (DiffuEraser, VideoPainter, and CLEAR) on our synthetic training set. SEDiT~\cite{hui2026sedit} and DeOverlay~\cite{miao2026deoverlay} are omitted from the quantitative comparison because their inference code was unavailable at the time of evaluation. We evaluate MiniMax-Remover only with its official checkpoint because the authors have not released training code.

\noindent
\textbf{Implementation Details.}
SmartEraser is built on Wan2.2-TI2V-5B~\cite{wan2025}. We train on 8 NVIDIA H200 GPUs with a global batch size of 24. We use AdamW~\cite{loshchilov2019decoupledweightdecayregularization} with a constant learning rate of $2 \times 10^{-5}$, $(\beta_1,\beta_2)=(0.9,0.999)$, a weight decay of $0.03$, and $\epsilon=10^{-10}$. During training, we set the noise scale $\beta$ to $0.005$. Our privileged mask curriculum learning strategy uses ground-truth overlay masks for the first 600 optimization steps and then follows a cosine schedule that reduces their probability of use to zero at step 4,200.  The 4,800-step curriculum allocates the final 600 steps to training without ground-truth masks. We fine-tune DiffuEraser, VideoPainter, and CLEAR on our synthetic training set for 48 hours each, using a single NVIDIA H200 GPU per method.

\begin{table*}[!t]
  \centering
  \caption{\textbf{Quantitative evaluation on SmartEraser-Bench synthetic videos.}
    $^{\spadesuit}$ denotes fine-tuning on our synthetic dataset. The best and second-best results are in \best{bold} and \second{underlined}, respectively.}
  \label{tab:1k}
  \resizebox{\textwidth}{!}{%
    \begin{tabular}{llcccccccc}
      \toprule
      & &
      \multicolumn{2}{c}{Overlay Removal} &
      \multicolumn{2}{c}{Background Preservation} &
      \multicolumn{3}{c}{Temporal Consistency} & \\
      \cmidrule(lr){3-4}
      \cmidrule(lr){5-6}
      \cmidrule(lr){7-9}
      Method & Venues
      & \shortstack{Masked\\PSNR $\uparrow$}
      & \shortstack{Masked\\LPIPS $\downarrow$}
      & \shortstack{Unmasked\\PSNR $\uparrow$}
      & \shortstack{Unmasked\\LPIPS $\downarrow$}
      & \shortstack{TC $\downarrow$\\$\times 10^3$}
      & \shortstack{TWE $\downarrow$\\$\times 10^3$}
      & \shortstack{Flow\\Mean $\downarrow$}
      & VFID $\downarrow$ \\
      \midrule
      DiffuEraser \cite{li2025diffueraser}
        & 2025-arXiv & 28.43 & 0.156 & 33.75 & \second{0.061}
        & \second{6.478} & 0.459 & 0.576 & \second{0.205} \\
      DiffuEraser$^{\spadesuit}$ \cite{li2025diffueraser}
        & 2025-arXiv & \second{28.97} & \second{0.151} & \second{34.13} & 0.063
        & 6.637 & 0.457 & 0.575 & 0.245 \\
      VideoPainter \cite{bian2025videopainter}
        & 2025-SIGGRAPH & 13.43 & 0.628 & 31.43 & 0.146
        & 10.226 & 1.142 & 0.622 & 2.350 \\
      VideoPainter$^{\spadesuit}$ \cite{bian2025videopainter}
        & 2025-SIGGRAPH & 14.82 & 0.573 & 31.37 & 0.138
        & 10.553 & 1.085 & 0.579 & 2.144 \\
      MiniMax-Remover \cite{zi2026minimax}
        & 2025-NeurIPS & 28.09 & 0.160 & 32.96 & \second{0.061}
        & 7.069 & \second{0.426} & \best{0.543} & 0.212 \\
      CLEAR \cite{he2026clear}
        & 2026-ICML & 12.66 & 0.713 & 22.10 & 0.207
        & 12.011 & 1.391 & 0.617 & 2.188 \\
      CLEAR$^{\spadesuit}$ \cite{he2026clear}
        & 2026-ICML & 21.20 & 0.184 & 21.30 & 0.163
        & 10.400 & 0.456 & 0.609 & 0.671 \\
      \midrule
      \rowcolor{oursrow}
      SmartEraser (Ours) & 2026
        & \best{34.38} & \best{0.074} & \best{36.59} & \best{0.045}
        & \best{6.105} & \best{0.409} & \second{0.555} & \best{0.076} \\
      \bottomrule
    \end{tabular}
  }
\end{table*}

\begin{table*}[!t]
  \centering
  \captionsetup{width=\textwidth}
  \caption{\textbf{Quantitative evaluation on SmartEraser-Bench real-world videos.}
    $^{\spadesuit}$ denotes fine-tuning on our synthetic dataset.
    The best and second-best results are in \best{bold} and
    \second{underlined}, respectively.}
  \label{tab:real_world}
  \resizebox{\textwidth}{!}{%
    \begin{tabular}{llccccccc}
      \toprule
      & &
      \multicolumn{2}{c}{Overlay Removal} &
      \multicolumn{2}{c}{Background Preservation} &
      \multicolumn{3}{c}{Temporal Consistency} \\
      \cmidrule(lr){3-4}
      \cmidrule(lr){5-6}
      \cmidrule(lr){7-9}
      Method & Venues
      & \shortstack{Success\\Rate $\uparrow$}
      & \shortstack{Visual\\Quality $\uparrow$}
      & \shortstack{Unmasked\\PSNR $\uparrow$}
      & \shortstack{Unmasked\\LPIPS $\downarrow$}
      & \shortstack{CLIP\\TC $\uparrow$}
      & \shortstack{Output-flow\\TWE $\times 10^3$$\downarrow$}
      & \shortstack{Flow\\Mean $\downarrow$} \\
      \midrule
      DiffuEraser~\cite{li2025diffueraser}
        & 2025-arXiv & 0.88 & 8.14 & \second{33.33} & \best{0.037}
        & \best{0.990} & 1.028 & 0.759 \\
      DiffuEraser$^{\spadesuit}$~\cite{li2025diffueraser}
        & 2025-arXiv & 0.86 & \second{8.21} & 33.32 & \best{0.037}
        & \second{0.989} & 1.032 & 0.760 \\
      VideoPainter~\cite{bian2025videopainter}
        & 2025-SIGGRAPH & 0.02 & 4.25 & 30.88 & 0.088
        & \best{0.990} & 0.902 & \second{0.708} \\
      VideoPainter$^{\spadesuit}$~\cite{bian2025videopainter}
        & 2025-SIGGRAPH & 0.01 & 3.94 & 30.83 & 0.088
        & \best{0.990} & \second{0.901} & \best{0.704} \\
    MiniMax-Remover~\cite{zi2026minimax}
        & 2025-NeurIPS & 0.89 & 8.14 & 31.89 & 0.046
        & \best{0.990} & 0.987 & 0.748 \\
      CLEAR~\cite{he2026clear}
        & 2026-ICML & 0.42 & 5.31 & 22.03 & 0.133
        & \second{0.989} & 1.150 & 0.772 \\
      CLEAR$^{\spadesuit}$~\cite{he2026clear}
        & 2026-ICML & \second{0.90} & 7.59 & 20.41 & 0.166
        & \second{0.989} & \best{0.863} & 0.778 \\
      \midrule
      \rowcolor{oursrow}
      SmartEraser (Ours)
        & 2026 & \best{0.94} & \best{8.60} & \best{35.92} & \second{0.039}
        & \best{0.990} & 0.997 & 0.751 \\
      \bottomrule
    \end{tabular}%
  }
\end{table*}

\begin{table}[t]
  \centering
  \caption{\textbf{Mask requirement and inference cost.} All methods process the same 121-frame video at a resolution of $720 \times 1280$ on a single NVIDIA H200 GPU. Reported runtimes measure inference only, excluding the time spent loading model weights, reading the input video, and saving the output video.}
  \label{tab:efficiency}
  \resizebox{\linewidth}{!}{%
  \begin{tabular}{lccc}
    \toprule
    Method & Mask-free
    & \shortstack{Inference\\steps}
    & \shortstack{Runtime\\(s/frame) $\downarrow$} \\
    \midrule
    DiffuEraser \cite{li2025diffueraser}     & \maskguided & 2  &0.49 \\
    VideoPainter \cite{bian2025videopainter} & \maskguided & 50 &4.81 \\
    MiniMax-Remover \cite{zi2026minimax}     & \maskguided & 12 &0.88 \\
    CLEAR \cite{he2026clear}                 & \maskfree   & 5  &0.57 \\
    \midrule
    \rowcolor{oursrow}
    SmartEraser (Ours)                       & \maskfree   & 1  &\best{0.08} \\
    \bottomrule
  \end{tabular}
  }
\end{table}

\subsection{Quantitative Evaluation}
\label{sec:quantitative_evaluation}

\paragraph{Synthetic videos.}
Table~\ref{tab:1k} shows that SmartEraser achieves the best values on seven of the eight reported metrics. Its largest gains occur within the overlay regions. Compared with fine-tuned DiffuEraser, the strongest baseline on both masked metrics, SmartEraser increases masked PSNR from 28.97 to 34.38\,dB and reduces masked LPIPS from 0.151 to 0.074, a 51.0\% reduction. These gains extend to the unmasked regions, where SmartEraser exceeds the best baseline PSNR by 2.46\,dB and lowers LPIPS from 0.061 to 0.045. The improvements in both regions indicate that the restored content and the surrounding background are closer to the clean target in both pixel and perceptual comparisons. SmartEraser also achieves the lowest TC and TWE, indicating better agreement with the target's frame-to-frame changes and lower residual error after alignment using target flow. Its VFID is 0.076, compared with 0.205 for the strongest baseline, a 62.9\% reduction in the distance between the restored and clean video feature distributions. Appendix~\ref{more_quantitative_results} presents quantitative comparisons for each of the 10 synthetic subsets.

\paragraph{Real-world videos.}
SmartEraser achieves the highest removal success rate (94\%) and visual quality score (8.60 out of 10) in Table~\ref{tab:real_world}. The margins over the best baselines are four percentage points in success rate and 0.39 points in visual quality. SmartEraser also achieves the highest unmasked PSNR (35.92\,dB), exceeding the next-best result by 2.59\,dB. DiffuEraser and its fine-tuned variant achieve the lowest unmasked LPIPS (0.037), followed by SmartEraser (0.039). The different rankings show that pixel-level agreement with the original background and perceptual similarity capture complementary aspects of preservation.

The temporal metrics provide a different view of the results. CLIP TC ranges from 0.989 to 0.990 at the reported precision, offering little separation among methods. Fine-tuned CLEAR achieves the lowest Output-flow TWE (0.863), and fine-tuned VideoPainter achieves the lowest Flow Mean (0.704). The latter nevertheless scores 3.94 in visual quality and succeeds on only 1\% of the clips. A persistent repair artifact can remain stable across adjacent frames, and Flow Mean measures the amount of apparent motion. These metrics are therefore most informative when read alongside removal and preservation scores.

\paragraph{Effect of fine-tuning.}
Fine-tuning improves masked PSNR and LPIPS for DiffuEraser, VideoPainter, and CLEAR on the synthetic split, with CLEAR showing the largest gains. Its masked PSNR rises from 12.66 to 21.20\,dB, and its masked LPIPS falls from 0.713 to 0.184. On real-world videos, fine-tuning raises CLEAR's success rate from 42\% to 90\% and its visual quality score from 5.31 to 7.59. At the same time, its unmasked PSNR decreases from 22.03 to 20.41\,dB and its unmasked LPIPS increases from 0.133 to 0.166. This trade-off shows that adapting a model to the overlay removal task can improve removal while reducing background fidelity. Improvements on the synthetic split also do not consistently carry over to real-world videos, as illustrated by VideoPainter's improved masked scores on synthetic clips and lower visual quality after fine-tuning on real clips.

\paragraph{Inference efficiency.}
As shown in Table~\ref{tab:efficiency}, SmartEraser removes overlays
without an input mask in a single inference step, achieving the lowest runtime of 0.08\,s/frame. On the same 121-frame clip, SmartEraser is approximately $6.1\times$ faster than DiffuEraser and $7.1\times$ faster than CLEAR. 

\begin{figure*}[!t]
  \centering
  \includegraphics[width=\textwidth]{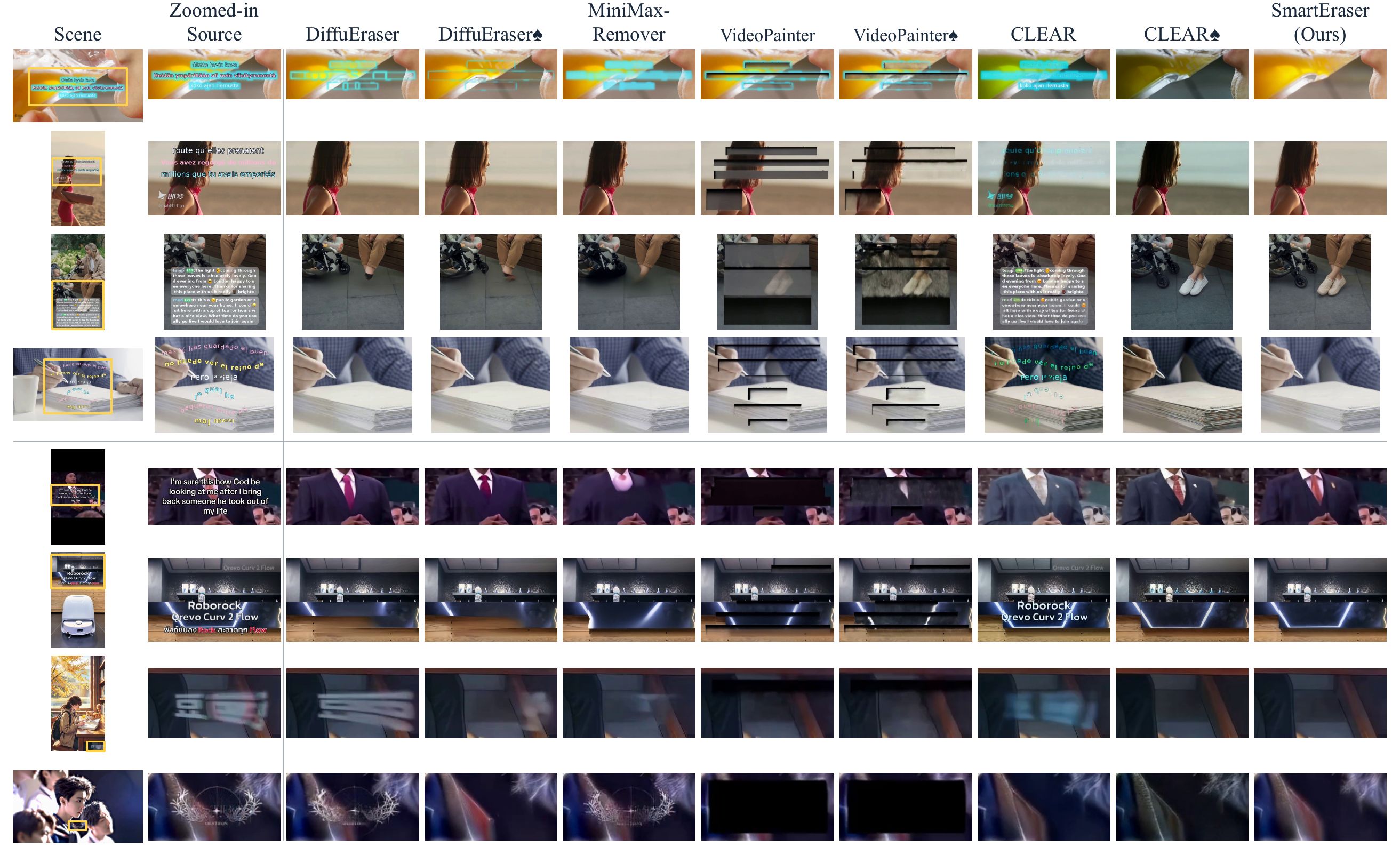}
  \caption{\textbf{Qualitative comparison on SmartEraser-Bench.} The top four rows show synthetic videos, and the bottom four rows show real-world videos. $^{\spadesuit}$ denotes fine-tuning on our synthetic training set.}
  \label{fig:qualitative_comparison}
\end{figure*}

\subsection{Qualitative Results}
\label{sec:qualitative_evaluation}
Fig.~\ref{fig:qualitative_comparison} presents qualitative comparisons on the synthetic and real-world splits of SmartEraser-Bench. In the synthetic examples, SmartEraser removes subtitle halos and dense live comment panels while preserving surface highlights and pavement seams. Several baselines leave residual glow or dark rectangular artifacts. In the real-world examples, SmartEraser removes multiline captions, product text, and a moving watermark while retaining scene details such as jacket lapels, diagonal light strips, and panel edges.

\begin{figure}[!t]
  \centering
  \includegraphics[width=\linewidth]{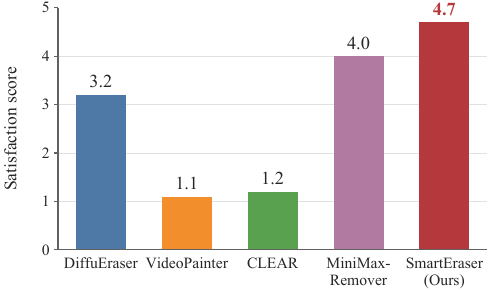}
  \caption{User satisfaction with video overlay removal.}
  \label{fig:user_study}
\end{figure}

\subsection{Ablation Study}
We conduct ablation studies to evaluate how OCRM and PMC contribute to one-step, mask-free video overlay removal. The experimental comparisons and their analysis are presented in Appendix~\ref{appendix_ablation}.

\subsection{User Study}
We conducted a user study with 28 participants, including researchers and students specializing in image and video editing. Participants evaluated overlay removal results for 20 video cases randomly sampled from SmartEraser-Bench. They rated their satisfaction on a scale with a maximum score of 5. As shown in Fig.~\ref{fig:user_study}, SmartEraser received the highest satisfaction score of 4.7.
\section{Conclusion}
We presented SmartEraser for one-step, mask-free video overlay removal. It combines Overlay-to-Clean Residual Matching with a Privileged Mask Curriculum that gradually withdraws mask guidance during training. We also introduced SmartEraser-Bench, comprising 1,000 synthetic video pairs across ten overlay categories and 100 real-world videos. With one-step, mask-free inference, SmartEraser performs comparably to multi-step mask-based methods and outperforms existing mask-free methods.

\section*{LLM Usage Statement}
During the preparation of this work, the authors used ChatGPT for grammar checking, language polishing, and rephrasing. After using this tool, the authors reviewed and edited the content as needed and take full responsibility for the content of the publication.

{
    \small
    \bibliographystyle{ieeenat_fullname}
    \bibliography{main}
}

\clearpage
\setcounter{page}{1}

\twocolumn[{%
  \centering
  {\Large\bfseries\thetitle\par}
  \vspace{0.5em}
  {\Large Supplementary Material\par}
  \vspace{1em}

  \begin{minipage}{\textwidth}
    \centering
    \includegraphics[width=\linewidth]{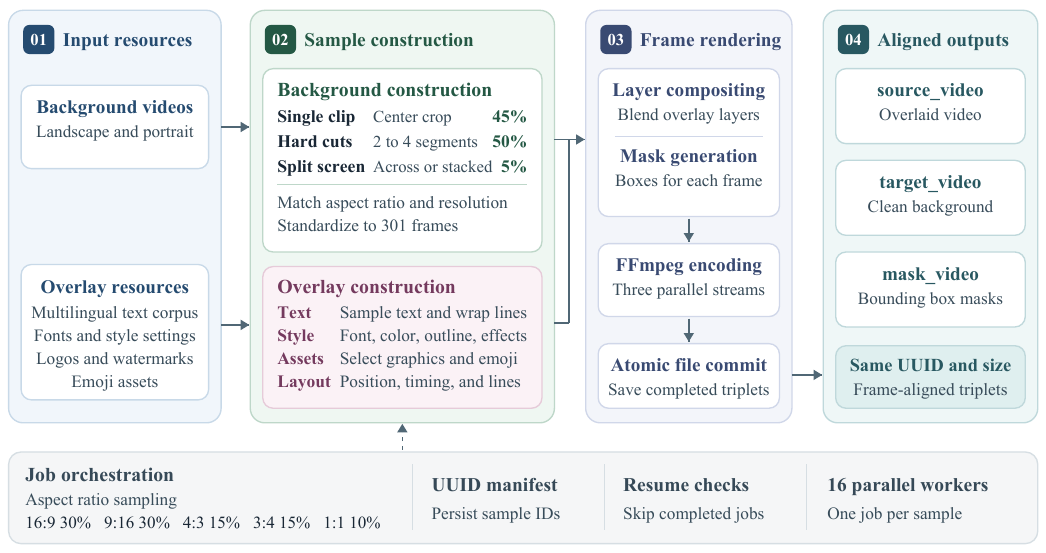}
    \captionof{figure}{Training data construction pipeline.}
    \label{training_data_pipeline}
  \end{minipage}\par
  \vspace{1em}
}]

\section{Dataset}
\subsection{Training Data}
\label{training_data}
As shown in Fig.~\ref{training_data_pipeline}, we constructed a training dataset of 50,000 samples for supervised video overlay removal. Each sample consists of three frame-aligned, 301-frame videos: an input video containing synthetic overlays (source video), the corresponding clean background video (target video), and a video of bounding-box masks indicating the overlay regions in each frame (mask video). By rendering the input from the clean background, the pipeline provides a corresponding target frame for every frame in which an overlay obscures the underlying content.

\textbf{Background construction.} We used both landscape and portrait videos as background material. For each sample, the output aspect ratio was drawn from a predefined distribution: 16:9 and 9:16 each had a sampling probability of 30\%, 4:3 and 3:4 each had a probability of 15\%, and 1:1 had a probability of 10\%. Background sequences were generated using three composition modes. In the single-clip mode (45\%), one clip was center-cropped. In the hard-cut mode (50\%), two to four segments were concatenated with abrupt transitions between adjacent segments. In the split-screen mode (5\%), clips were arranged side by side or stacked vertically within the same frame. The resulting background was formatted to match the selected aspect ratio and output resolution and standardized to 301 frames.

\textbf{Overlay construction.} Overlay resources included a multilingual text corpus, fonts and style settings, logos, watermarks, and emoji assets. Text was sampled from the corpus and wrapped into lines. Their appearance was specified through the choice of font, color, outline, and visual effects. Graphical overlays were assembled from selected logo, watermark, and emoji assets. The layout specified where each overlay appeared in the frame, when it appeared in the sequence, and how text was arranged across lines.

\begin{table*}[!t]
  \centering
  \small
  \caption{Synthetic video categories in SmartEraser-Bench.
    Each category contains 100 videos, for a total of 1,000 videos.}
  \label{tab:smarteraser-bench-synthetic}
  \setlength{\tabcolsep}{3pt}
  \renewcommand{\arraystretch}{1}
  \newlength{\SEdescriptionwidth}
  \setlength{\SEdescriptionwidth}{\dimexpr\textwidth-.058\textwidth-.250\textwidth-4\tabcolsep\relax}
  \newlength{\SEframeheight}
  \setlength{\SEframeheight}{\dimexpr(\textheight-295pt)/10\relax}
  \ifdim\SEframeheight>40pt
    \setlength{\SEframeheight}{40pt}
  \fi
  \newcommand{\SEshot}[2]{%
    \parbox[c][\SEframeheight][c]{.242\linewidth}{\centering
      \includegraphics[width=\linewidth,height=\SEframeheight,keepaspectratio]%
        {figures/synthetic_examples/#1_#2.jpg}}}
  \newcommand{\SEclip}[1]{%
    \begin{minipage}[t]{.488\textwidth}
      \noindent\SEshot{#1}{1}\hfill
      \SEshot{#1}{2}\hfill
      \SEshot{#1}{3}\hfill
      \SEshot{#1}{4}%
    \end{minipage}}
  \newcommand{\SEbody}{%
  \footnotesize
  \begin{tabular}{@{}>{\centering\arraybackslash}p{.058\textwidth}
                     >{\raggedright\arraybackslash}p{.250\textwidth}
                     >{\raggedright\arraybackslash}p{\SEdescriptionwidth}@{}}
    \toprule
    \textbf{No.} & \textbf{Category} & \textbf{Description} \\
    \midrule
    1 & \textbf{Standard / Rotated} & Subtitles appear horizontally or at a range of rotation angles. \\
    \multicolumn{3}{@{}p{\textwidth}@{}}{%
      \SEclip{01_a}\hfill
      \SEclip{01_b}%
    } \\
    \midrule
    2 & \textbf{Emojis \& Stickers} & Emojis embedded in subtitles and separate stickers vary across frames. \\
    \multicolumn{3}{@{}p{\textwidth}@{}}{%
      \SEclip{02_a}\hfill
      \SEclip{02_b}%
    } \\
    \midrule
    3 & \textbf{Glowing Subtitles} & Luminous halos extend beyond the subtitle glyphs. \\
    \multicolumn{3}{@{}p{\textwidth}@{}}{%
      \SEclip{03_a}\hfill
      \SEclip{03_b}%
    } \\
    \midrule
    4 & \textbf{Full-Screen Text} & Text fragments are distributed across the frame. \\
    \multicolumn{3}{@{}p{\textwidth}@{}}{%
      \SEclip{04_a}\hfill
      \SEclip{04_b}%
    } \\
    \midrule
    5 & \textbf{Comment Cards} & Graphical cards display comments alongside avatars and usernames. \\
    \multicolumn{3}{@{}p{\textwidth}@{}}{%
      \SEclip{05_a}\hfill
      \SEclip{05_b}%
    } \\
    \midrule
    6 & \textbf{Transition Blur} & Subtitle text blurs as the video transitions between scenes. \\
    \multicolumn{3}{@{}p{\textwidth}@{}}{%
      \SEclip{06_a}\hfill
      \SEclip{06_b}%
    } \\
    \midrule
    7 & \textbf{Live Chat} & Viewer messages appear and update in a stacked chat feed. \\
    \multicolumn{3}{@{}p{\textwidth}@{}}{%
      \SEclip{07_a}\hfill
      \SEclip{07_b}%
    } \\
    \midrule
    8 & \textbf{Dynamic Watermarks} & Platform watermarks change position or appearance over time. \\
    \multicolumn{3}{@{}p{\textwidth}@{}}{%
      \SEclip{08_a}\hfill
      \SEclip{08_b}%
    } \\
    \midrule
    9 & \textbf{Translucent Subtitles} & The underlying scene remains visible through the subtitle glyphs. \\
    \multicolumn{3}{@{}p{\textwidth}@{}}{%
      \SEclip{09_a}\hfill
      \SEclip{09_b}%
    } \\
    \midrule
    10 & \textbf{Curved \& Animated} & Curved subtitles change shape or placement over time. \\
    \multicolumn{3}{@{}p{\textwidth}@{}}{%
      \SEclip{10_a}\hfill
      \SEclip{10_b}%
    } \\
    \bottomrule
  \end{tabular}}
  \newsavebox{\SEtablebox}
  \sbox{\SEtablebox}{\SEbody}
  \newlength{\SEexcess}
  \setlength{\SEexcess}{%
    \dimexpr\ht\SEtablebox+\dp\SEtablebox-\textheight+56pt\relax}
  \ifdim\SEexcess>0pt
    \addtolength{\SEframeheight}{\dimexpr0pt-(\SEexcess+2pt)/10\relax}
    \sbox{\SEtablebox}{\SEbody}
  \fi
  \usebox{\SEtablebox}
\end{table*}

\textbf{Frame compositing and mask generation.} At each frame index, the active overlay layers were blended with the corresponding clean background frame to produce a source frame. The clean frame at the same index served as the target, and the bounding boxes of the overlays present at that index were used to generate the corresponding mask frame. The three streams were encoded in parallel using FFmpeg. All videos in a triplet shared the same spatial dimensions, and frame count, preserving the correspondence among the input, target, and mask at every frame index.

\subsection{SmartEraser-Bench Synthetic Videos}
\label{testing_data}
The synthetic split of SmartEraser-Bench comprises 1,000 videos, with 100 videos in each of 10 overlay categories. As shown in Table~\ref{tab:smarteraser-bench-synthetic}, the categories capture variations in subtitle orientation, visual effects, opacity, and spatial coverage, and include structured overlays such as comment cards and live chat feeds. For each category, we show four chronologically ordered frames from each of two videos. These sequences illustrate temporal effects such as subtitle blur during scene transitions, updates to live chat messages, and changes in the positions of platform watermarks. The benchmark evaluates whether removal methods can recover background details beneath these overlays while maintaining frame-to-frame consistency as the overlays change.

\section{Evaluation Metrics}
\label{app:metrics}

\paragraph{Notation and aggregation.}
Let $X_t$, $\widehat{Y}_t$, and $Y_t$ denote the aligned input, prediction,
and clean target at frame $t$, respectively, with $t=1,\ldots,T$.
RGB values are in $[0,1]$ unless specified otherwise. We binarize each
decoded grayscale mask at an intensity greater than 127 on the 8-bit
scale. For a frame with pixel grid $\Omega$, the masked region is
$\mathcal{M}_t$ and the unmasked region is $\Omega\setminus\mathcal{M}_t$.
We average video-level scores across the evaluated clips. VFID is
computed separately from all feature pairs in the relevant split or
subset. The tables scale TC, TWE, and Output-flow TWE by $10^3$.

\paragraph{Masked and unmasked PSNR.}
Let $R_t$ denote the region being evaluated and let $B_t$ be the reference:
$Y_t$ for synthetic clips and $X_t$ for real-world clips. We pool RGB
squared errors over all pixels in the selected region across the clip,
then convert the resulting MSE to decibels:
\begin{align}
  \mathrm{MSE}_{R}
    &= \frac{\sum_{t=1}^{T}\sum_{p\in R_t}
       \lVert \widehat{Y}_t(p)-B_t(p)\rVert_2^2}
       {3\sum_{t=1}^{T}|R_t|},\\
  \mathrm{PSNR}_{R}
    &= -10\log_{10}\!\left(\mathrm{MSE}_{R}\right).
\end{align}
An empty region across the entire clip yields an undefined score. An exact match yields $+\infty$. We evaluate both regions on synthetic clips and only the unmasked region on real-world clips.

\paragraph{Masked and unmasked LPIPS.}
We use LPIPS v0.1~\cite{zhang2018unreasonable} with an AlexNet backbone~\cite{krizhevsky2017imagenet}, with image inputs mapped to $[-1,1]$.
For synthetic clips we compute $D_t$ from $\widehat{Y}_t$ and $Y_t$.
For real-world clips, we first restore the original input inside the mask
when forming the frame used for unmasked LPIPS:
\[
  \widetilde{Y}_t=(1-M_t)\odot\widehat{Y}_t+M_t\odot X_t,
\]
where $M_t$ is the binary indicator of $\mathcal{M}_t$. We then compute
$D_t$ from $\widetilde{Y}_t$ and $X_t$. Let $w_t$ be the binary indicator
of $R_t$, area-resampled if needed to the resolution of $D_t$, and let
$\mathcal{A}=\{t:\sum_p w_t(p)>0\}$. We average spatially within each
valid frame and then give every valid frame equal weight:
\[
  \mathrm{LPIPS}_{R}
  =\frac{1}{|\mathcal A|}\sum_{t\in\mathcal A}
      \frac{\sum_p w_t(p)D_t(p)}{\sum_p w_t(p)}.
\]
We report no score when $\mathcal{A}$ is empty.

\paragraph{Temporal Change Error (TC).}
We use the full-frame form of the Temporal Change Error
in MFVSR~\cite{huang2026mfvsr}. On synthetic clips, it compares
consecutive RGB differences in the prediction and clean target at the
original resolution:
\[
  \begin{aligned}
    \mathrm{TC}
    &= \frac{1}{3(T-1)|\Omega|}
       \sum_{t=2}^{T}\sum_{p\in\Omega} \\
    &\quad \Bigl\lVert
      [\widehat{Y}_t(p)-\widehat{Y}_{t-1}(p)]
      -[Y_t(p)-Y_{t-1}(p)]
    \Bigr\rVert_1 .
  \end{aligned}
\]

\paragraph{TWE and Output-flow TWE.}
We assess motion-compensated consistency using temporal warping
error~\cite{lai2018learning}. We preserve the aspect ratio, cap the
longer side at 320 pixels, and estimate dense Farneb\"ack
flow~\cite{farneback2003two} on grayscale frames. The OpenCV parameters are
\texttt{pyr\_scale=0.5}, \texttt{levels=3}, \texttt{winsize=15},
\texttt{iterations=3}, \texttt{poly\_n=5}, \texttt{poly\_sigma=1.2}, and
\texttt{flags=0}. Let $b_t$ denote backward flow from frame $t$ to
$t-1$ on the resized grid $\Omega'$. Let
$\mathcal{V}_t=\{p\in\Omega':p+b_t(p)\in[0,W'-1]\times[0,H'-1]\}$,
where $W'$ and $H'$ are the resized dimensions and pixel coordinates
start at zero. We bilinearly sample the preceding
predicted frame at $p+b_t(p)$ and compute
\begin{align}
 e_t &=
 \frac{1}{3|\mathcal{V}_t|}\sum_{p\in\mathcal{V}_t}
 \bigl\|\widehat{Y}_t(p)-\widehat{Y}_{t-1}(p+b_t(p))\bigr\|_2^2,\\
 \mathrm{TWE} &= \frac{1}{|\mathcal{P}|}\sum_{t\in\mathcal{P}}e_t,
 \qquad \mathcal{P}=\{t\in\{2,\ldots,T\}:|\mathcal{V}_t|>0\}.
\end{align}
Here the prediction frames in $e_t$ are the resized RGB frames.
For synthetic clips, we estimate $b_t$ on the clean target.
For real-world clips, we estimate $b_t$ on the prediction and call the
result \emph{Output-flow TWE}.

\paragraph{Flow Mean.}
On both splits, let $u_t(p)$ be the forward Farneb\"ack flow from
predicted frame $t$ to $t+1$, using the same resolution and flow parameters
as above. We average its Euclidean magnitude over the full video:
\[
  \mathrm{Flow\ Mean}=
  \frac{1}{(T-1)|\Omega'|}\sum_{t=1}^{T-1}
  \sum_{p\in\Omega'}\lVert u_t(p)\rVert_2 .
\]
The units are pixels per frame at the evaluation resolution. Flow Mean
describes the amount of apparent motion, so we interpret it together
with the fidelity and consistency metrics.

\begin{figure*}[!t]
  \centering
  \includegraphics[width=\textwidth]{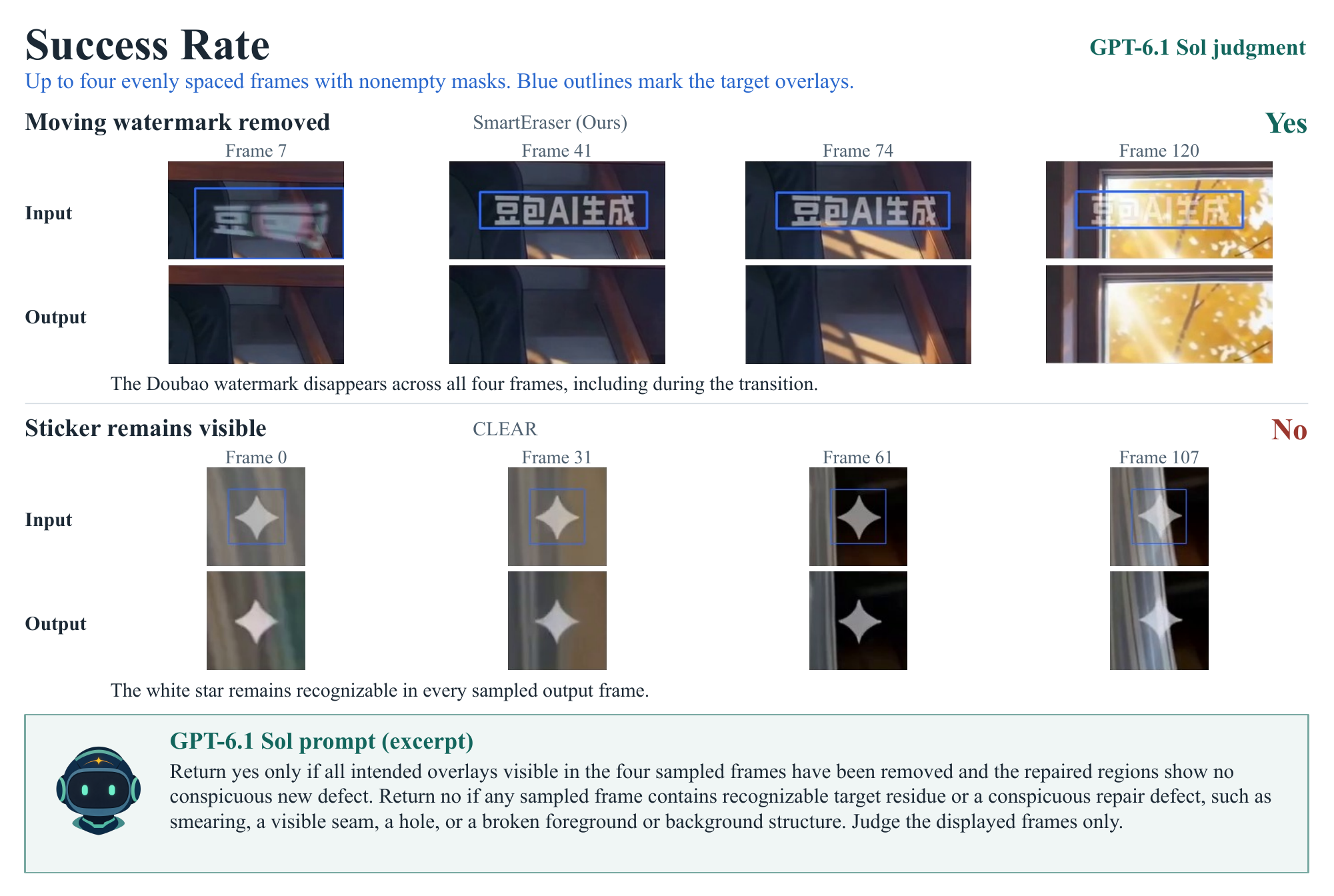}
  \caption{\textbf{Success Rate examples.} The evaluator assigns Yes to SmartEraser when the moving watermark is removed and No to CLEAR when the target sticker remains visible.}
  \label{fig:success_rate_examples}
\end{figure*}

\paragraph{CLIP TC.}
On the real-world split, we encode each full output frame using the
CLIP ViT-B/32 image encoder~\cite{radford2021learning}. We resize the
longer side to 224 pixels, preserve the aspect ratio, and pad with
replicated edge pixels to $224\times224$ before applying the CLIP
normalization. Let $\bar z_t$ be the $\ell_2$-normalized image embedding.
CLIP TC averages the cosine similarity of adjacent embeddings:
\[
  \mathrm{CLIP\ TC}=\frac{1}{T-1}\sum_{t=2}^{T}
  \bar z_{t-1}^{\mathsf T}\bar z_t.
\]

\paragraph{VFID.}
For each synthetic prediction and its clean target, we encode the
entire clip with RGB I3D~\cite{carreira2017quo}. We preserve the
aspect ratio, cap the longer side at 320 pixels, and map RGB to $[-1,1]$.
Spatiotemporal pooling of the final feature map before the classifier
yields one 1024-dimensional vector per clip. Let
$(\mu_{\mathrm{p}},\Sigma_{\mathrm{p}})$ and
$(\mu_{\mathrm{g}},\Sigma_{\mathrm{g}})$ denote the sample means and
covariances of the prediction and clean target features. We compute
VFID~\cite{wang2018video} as
\begin{align}
  \mathrm{VFID}
  &=\lVert\mu_{\mathrm{p}}-\mu_{\mathrm{g}}\rVert_2^2\\
  &\quad+\operatorname{tr}\!\left(
    \Sigma_{\mathrm{p}}+\Sigma_{\mathrm{g}}
    -2\left(\Sigma_{\mathrm{p}}^{1/2}\Sigma_{\mathrm{g}}
             \Sigma_{\mathrm{p}}^{1/2}\right)^{1/2}\right).
\end{align}
The result is a single distance for the full synthetic split or a
category subset. The real-world split has no aligned clean clips for
this comparison.

\begin{figure*}[!t]
  \centering
  \includegraphics[width=\textwidth]{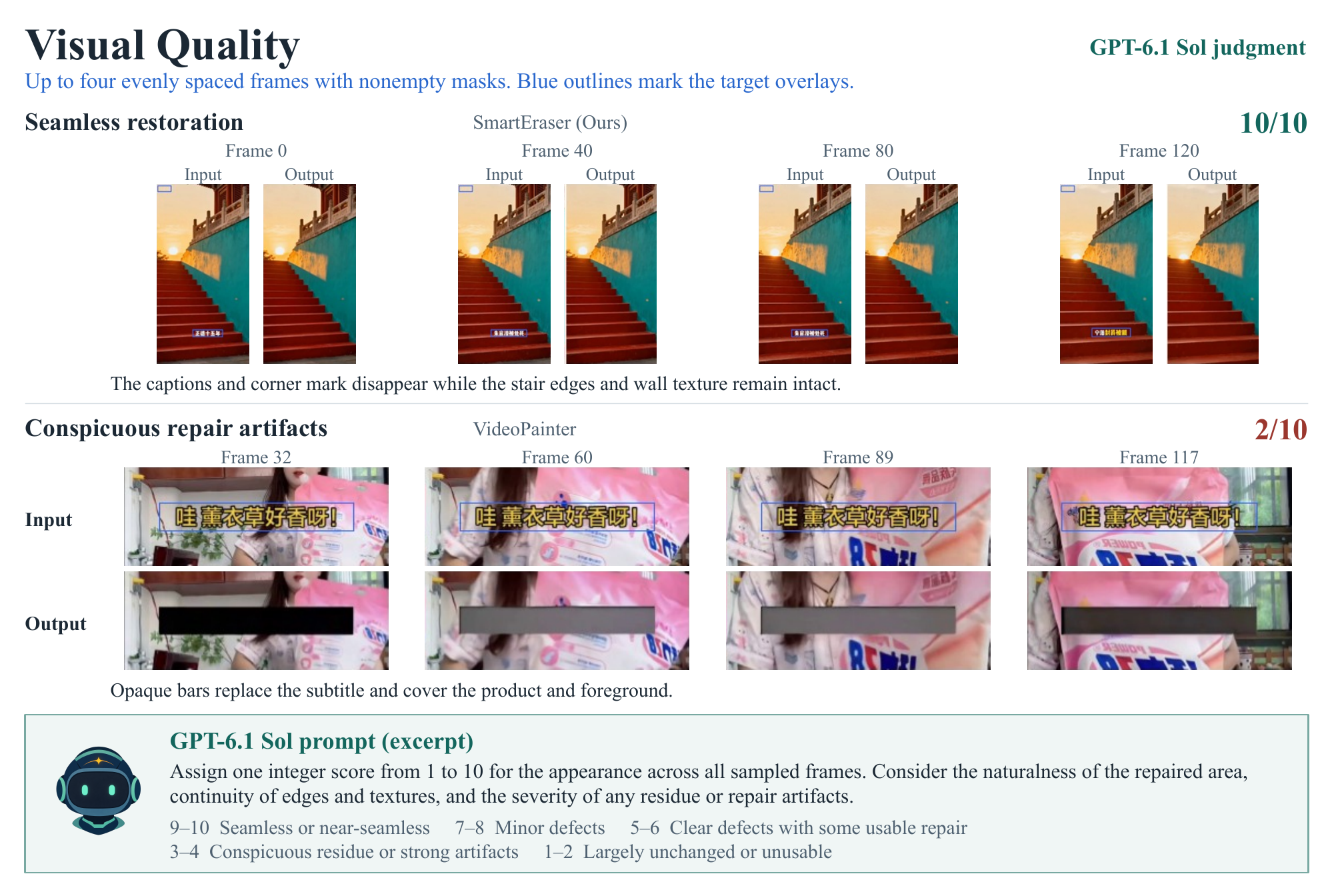}
  \caption{\textbf{Visual Quality examples.} The evaluator assigns
  10/10 to SmartEraser for seamless restoration and 2/10 to VideoPainter
  for conspicuous bars that obscure scene content.}
  \label{fig:visual_quality_examples}
\end{figure*}

\paragraph{Success Rate and Visual Quality.}
We sample up to four evenly spaced positions from the ordered set of
frames with a nonempty mask, including its first and last frames.
GPT-6.1 Sol~\cite{openai2026gpt61sol} sees aligned crops of the input
and prediction, with the target regions outlined on the input.
It assigns a success label only when all target overlays visible in the
sampled frames are removed and the repaired regions show no conspicuous
defects. Recognizable residue or a conspicuous defect in any sampled
frame leads to a failure label. The scorer treats genuine scene text
and graphics as part of the scene. The model also assigns one integer quality score from 1 to 10
per video based on how natural the repaired area looks, the continuity of
edges and textures, and the severity of residue or artifacts.
For $N$ visually scored videos, let $s_i\in\{0,1\}$ be the success label
and $q_i\in\{1,\ldots,10\}$ the quality score:
\[
  \mathrm{Success\ Rate}=\frac{1}{N}\sum_{i=1}^{N}s_i,\qquad
  \mathrm{Visual\ Quality}=\frac{1}{N}\sum_{i=1}^{N}q_i.
\]
Fig.~\ref{fig:success_rate_examples} and~\ref{fig:visual_quality_examples}
show the scoring rules and judgments.


\begin{table*}[!t]
  \centering
  \caption{\textbf{Ablation on mask training strategies} All modes use one-step mask-free inference on
    the same 1,000 synthetic videos of SmartEraser-Bench. The best and second-best
    results in each column are shown in \textbf{bold} and
    \underline{underlined}, respectively.}
  \label{tab:mask_mode_ablation}
  \resizebox{\textwidth}{!}{%
  \tabcolsep=10 pt
    \begin{tabular}{lcccccccc}
      \toprule
      & \multicolumn{2}{c}{Overlay Removal}
      & \multicolumn{2}{c}{Background Preservation}
      & \multicolumn{3}{c}{Temporal Consistency} & \\
      \cmidrule(lr){2-3}
      \cmidrule(lr){4-5}
      \cmidrule(lr){6-8}
      Training mode
      & \shortstack{Masked\\PSNR $\uparrow$}
      & \shortstack{Masked\\LPIPS $\downarrow$}
      & \shortstack{Unmasked\\PSNR $\uparrow$}
      & \shortstack{Unmasked\\LPIPS $\downarrow$}
      & \shortstack{TC $\downarrow$\\$\times 10^3$}
      & \shortstack{TWE $\downarrow$\\$\times 10^3$}
      & \shortstack{Flow\\Mean $\downarrow$}
      & VFID $\downarrow$ \\
      \midrule
      Mode~1
        & 29.90 & 0.092 & 35.47 & 0.047
        & 6.204 & 0.4166 & 0.5538 & 0.101 \\
      Mode~2
        & 31.53 & 0.086 & \textbf{36.78} & \underline{0.046}
        & 6.217 & 0.4051 & \textbf{0.5507} & 0.092 \\
      Mode~3
        & \underline{32.63} & \underline{0.084}
        & 36.27 & \underline{0.046}
        & \underline{6.160} & \textbf{0.4041}
        & \underline{0.5515} & \underline{0.084} \\
      Mode~4
        & 31.20 & 0.088 & 35.98 & \underline{0.046}
        & 6.173 & \underline{0.4042} & 0.5526 & 0.090 \\
      \rowcolor{oursrow}
      Mode~5 (SmartEraser)
        & \textbf{34.38} & \textbf{0.074}
        & \underline{36.59} & \textbf{0.045}
        & \textbf{6.105} & 0.4089 & 0.5551 & \textbf{0.076} \\
      \bottomrule
    \end{tabular}%
  }
\end{table*}

\section{Ablation Study}
\label{appendix_ablation}
\begin{figure}[!t]
  \centering
  \includegraphics[width=\columnwidth]{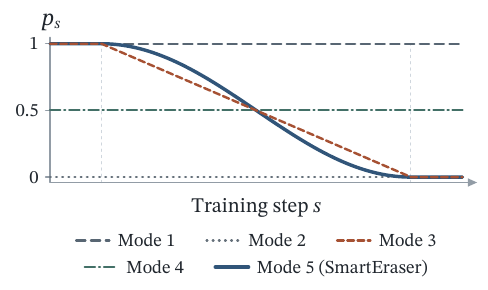}
  \caption{\textbf{Mask sampling strategies during training.} At training step $s$, $p_s$ is the probability of conditioning the model on the ground-truth overlay mask. With probability $1-p_s$, the model receives a null mask whose entries are all $0.5$.}
  \label{fig:mask_training_schedules}
\end{figure}

\paragraph{Mask Training Strategies.} We compare the five mask sampling strategies in Fig.~\ref{fig:mask_training_schedules} using one-step mask-free inference. The cosine PMC schedule retains a higher probability of GT-mask conditioning than linear decay in the first half of the transition and a lower probability in the second half. Early GT masks provide explicit spatial guidance for learning where to remove overlays and where to preserve scene content. As training progresses, more frequent null-mask conditioning gives the model greater practice in locating and removing overlays from the input video alone, matching the inference setting. The cosine schedule also has zero slope at both ends of the transition, connecting the GT-only and null-only phases smoothly. This progression may help explain why Mode~5 performs best overall in Tab.~\ref{tab:mask_mode_ablation}, leading on five of the eight metrics.

\begin{table*}[!t]
  \centering
  \caption{\textbf{Ablation study of Overlay-to-Clean Residual Matching on SmartEraser-Bench.} The full training grid is $\mathcal{T}=\{250,500,750,1000\}$. Exp.~1 trains exclusively at $t=1000$. Exp.~2 removes the explicit overlay latent $z_1$ from the conditioning channels at internal timesteps while retaining it at $t=1000$. The best and second-best results are in \best{bold} and \second{underlined}, respectively.}
  \label{tab:ocrm_ablation}
  \resizebox{\textwidth}{!}{%
    \setlength{\tabcolsep}{5pt}
    \begin{tabular}{lcccccccccc}
      \toprule
      & & & \multicolumn{2}{c}{Overlay Removal} & \multicolumn{2}{c}{Background Preservation} & \multicolumn{3}{c}{Temporal Consistency} & \\
      \cmidrule(lr){4-5}
      \cmidrule(lr){6-7}
      \cmidrule(lr){8-10}
      Variant & \shortstack{Training\\timesteps} & \shortstack{Internal\\overlay condition} & \shortstack{Masked\\PSNR $\uparrow$} & \shortstack{Masked\\LPIPS $\downarrow$} & \shortstack{Unmasked\\PSNR $\uparrow$} & \shortstack{Unmasked\\LPIPS $\downarrow$} & \shortstack{TC $\downarrow$\\$\times 10^3$} & \shortstack{TWE $\downarrow$\\$\times 10^3$} & \shortstack{Flow\\Mean $\downarrow$} & VFID $\downarrow$ \\
      \midrule
      Exp.~1: endpoint only & $\{1000\}$ & N/A & \underline{32.61} & \underline{0.082} & 35.96 & \underline{0.047} & \underline{6.115} & \underline{0.411} & \underline{0.556} & 0.094 \\
      Exp.~2: no internal $z_1$ & $\mathcal{T}$ & Removed & 31.81 & 0.083 & \underline{36.37} & \textbf{0.045} & 6.123 & 0.417 & \textbf{0.555} & \underline{0.085} \\
      \midrule
      \rowcolor{oursrow}
      Full OCRM (SmartEraser) & $\mathcal{T}$ & Retained & \textbf{34.38} & \textbf{0.074} & \textbf{36.59} & \textbf{0.045} & \textbf{6.105} & \textbf{0.409} & \textbf{0.555} & \textbf{0.076} \\
      \bottomrule
    \end{tabular}%
  }
\end{table*}

\paragraph{Effectiveness of OCRM.}
We evaluate OCRM through two ablations that examine the contributions of supervision at internal timesteps and explicit overlay conditioning. As shown in Table~\ref{tab:ocrm_ablation}, Exp.~1 restricts training to $\mathcal{T}=\{1000\}$, allocating all training samples to the deployment timestep. Exp.~2 retains the four-point grid $\mathcal{T}=\{250,500,750,1000\}$ but removes the overlay latent $z_1$ from the conditioning channels at internal timesteps. 

Our OCRM objective supervises a common removal residual across different intermediate latent states. Given the overlay latent $z_1=\mathcal{E}(V_1)$, the clean latent $z_0=\mathcal{E}(V_0)$, and the text condition $c=\mathrm{T5}(P)$, we construct
\begin{equation}
  \begin{aligned}
    \eta_t &= \beta\sqrt{\sigma_t(1-\sigma_t)}\,\epsilon, \qquad \epsilon\sim\mathcal{N}(0,I), \\
    z_t &= \sigma_t z_1+(1-\sigma_t)z_0+\eta_t, \\
    r &= z_1-z_0.
  \end{aligned}
  \label{eq:ocrm_state_ablation}
\end{equation}
Substituting the residual definition gives
\begin{equation}
  z_t=z_0+\sigma_t r+\eta_t.
  \label{eq:ocrm_residual_interpolation}
\end{equation}
Thus, the deterministic component of $z_t$ contains different amplitudes of the same removal residual, while the supervised target remains $r$ at every timestep. We train the shared predictor using
\begin{equation}
  \widehat r=v_\theta(z_t,t,[M_s,z_1],c), \qquad
  \mathcal{L}_t=\mathbb{E}\!\left[\operatorname{MSE}(\widehat r,r)\mid t\right],
  \label{eq:ocrm_timestep_loss}
\end{equation}
where $M_s$ follows our cosine mask sampling schedule. Because timesteps are sampled uniformly, the full objective is
\begin{equation}
  \mathcal{L}_{\mathrm{OCRM}}=\frac{1}{4}\mathcal{L}_{1000}+\frac{1}{4}\sum_{t\in\{250,500,750\}}\mathcal{L}_t.
  \label{eq:ocrm_multitimestep_objective}
\end{equation}
The internal terms update the same parameters used at deployment, allowing supervision from intermediate states to shape the final endpoint predictor. This formulation gives internal supervision a direct restoration interpretation. For every sampled timestep, we have
\begin{equation}
  \operatorname{MSE}(\widehat r,r)=\operatorname{MSE}(z_1-\widehat r,z_0).
  \label{eq:ocrm_reconstruction_equivalence}
\end{equation}
Matching the complete removal residual therefore also supervises reconstruction of the same clean latent from every training state. The model encounters varying residual amplitudes and noise perturbations while learning a consistent clean reconstruction. This couples the tasks at different timesteps and encourages a shared representation of the overlay-to-clean transformation. At the deployment endpoint, $\sigma_{1000}=1$ and $z_{1000}=z_1$, so the learned transformation is applied directly to the overlay latent.

\begin{table*}[!t]
  \centering
  \caption{\textbf{Ablation on the number of inference steps.}
    All settings use the same 1,000 synthetic videos of SmartEraser-Bench.
    The best and second-best results are in \best{bold} and \second{underlined}, respectively.}
  \label{tab:step_ablation}
  \resizebox{\textwidth}{!}{%
  \tabcolsep=15 pt
    \begin{tabular}{ccccccccc}
      \toprule
      & \multicolumn{2}{c}{Overlay Removal}
      & \multicolumn{2}{c}{Background Preservation}
      & \multicolumn{3}{c}{Temporal Consistency} & \\
      \cmidrule(lr){2-3}
      \cmidrule(lr){4-5}
      \cmidrule(lr){6-8}
      Steps
      & \shortstack{Masked\\PSNR $\uparrow$}
      & \shortstack{Masked\\LPIPS $\downarrow$}
      & \shortstack{Unmasked\\PSNR $\uparrow$}
      & \shortstack{Unmasked\\LPIPS $\downarrow$}
      & \shortstack{TC $\downarrow$\\$\times 10^3$}
      & \shortstack{TWE $\downarrow$\\$\times 10^3$}
      & \shortstack{Flow\\Mean $\downarrow$}
      & VFID $\downarrow$ \\
      \midrule
      \rowcolor{oursrow}
      1 & 34.38 & \textbf{0.074} & 36.59 & \textbf{0.045}
        & \textbf{6.105} & \textbf{0.409} & \textbf{0.555} & \underline{0.076} \\
      2 & \underline{34.57} & \textbf{0.074} & \underline{36.73} & \textbf{0.045}
        & \underline{6.109} & \underline{0.411} & \underline{0.557} & \textbf{0.075} \\
      3 & 33.57 & 0.087 & 36.67 & \underline{0.049}
        & 6.464 & 0.449 & \underline{0.557} & 0.108 \\
      4 & \textbf{34.58} & \underline{0.077} & \textbf{36.89} & \textbf{0.045}
        & 6.173 & 0.418 & \underline{0.557} & 0.082 \\
      \bottomrule
    \end{tabular}%
  }
\end{table*}

\begin{table*}[!t]
  \centering
  \caption{\textbf{Quantitative evaluation on the \red{Standard and Rotated Text} subset of SmartEraser-Bench synthetic videos.}
    $^{\spadesuit}$ denotes fine-tuning on our synthetic dataset. The best and second-best results are in \best{bold} and \second{underlined}, respectively.}
  \label{tab:normal_and_rotated_text}
  \resizebox{\textwidth}{!}{%
    \begin{tabular}{llcccccccc}
      \toprule
      & &
      \multicolumn{2}{c}{Overlay Removal} &
      \multicolumn{2}{c}{Background Preservation} &
      \multicolumn{3}{c}{Temporal Consistency} & \\
      \cmidrule(lr){3-4}
      \cmidrule(lr){5-6}
      \cmidrule(lr){7-9}
      Method & Venues
      & \shortstack{Masked\\PSNR $\uparrow$}
      & \shortstack{Masked\\LPIPS $\downarrow$}
      & \shortstack{Unmasked\\PSNR $\uparrow$}
      & \shortstack{Unmasked\\LPIPS $\downarrow$}
      & \shortstack{TC $\downarrow$\\$\times 10^3$}
      & \shortstack{TWE $\downarrow$\\$\times 10^3$}
      & \shortstack{Flow\\Mean $\downarrow$}
      & VFID $\downarrow$ \\
      \midrule
      DiffuEraser \cite{li2025diffueraser} & 2025-arXiv & 29.69 & 0.114 & \second{34.89} & 0.058 & \second{5.885} & 0.172 & 0.441 & 0.323 \\
      DiffuEraser$^{\spadesuit}$ \cite{li2025diffueraser} & 2025-arXiv & 29.25 & 0.127 & \second{34.89} & 0.060 & 6.110 & 0.179 & 0.438 & 0.420 \\
      VideoPainter \cite{bian2025videopainter} & 2025-SIGGRAPH & 13.67 & 0.622 & 32.60 & 0.130 & 9.801 & 1.037 & 0.470 & 3.554 \\
      VideoPainter$^{\spadesuit}$ \cite{bian2025videopainter} & 2025-SIGGRAPH & 15.72 & 0.534 & 32.65 & 0.124 & 10.118 & 1.069 & 0.464 & 3.343 \\
      MiniMax-Remover \cite{zi2026minimax} & 2025-NeurIPS & \second{30.61} & \second{0.102} & 34.46 & \second{0.055} & 6.034 & \second{0.151} & \best{0.409} & \second{0.312} \\
      CLEAR \cite{he2026clear} & 2026-ICML & 12.01 & 0.713 & 21.76 & 0.201 & 11.387 & 1.709 & 0.456 & 4.427 \\
      CLEAR$^{\spadesuit}$ \cite{he2026clear} & 2026-ICML & 20.85 & 0.181 & 20.98 & 0.168 & 8.904 & 0.223 & 0.469 & 1.874 \\
      \midrule
      \rowcolor{oursrow}
      SmartEraser (Ours) & 2026 & \best{35.16} & \best{0.065} & \best{37.06} & \best{0.044} & \best{5.321} & \best{0.135} & \second{0.424} & \best{0.173} \\
      \bottomrule
    \end{tabular}
  }
\end{table*}

Full OCRM improves masked PSNR over endpoint-only training by $1.77$~dB, from $32.61$ to $34.38$~dB, while reducing masked LPIPS from $0.082$ to $0.074$. These gains demonstrate the effectiveness of distributing residual supervision across training timesteps. Although the endpoint receives one quarter of the training samples in full OCRM, the shared predictor achieves better endpoint restoration. The additional intermediate states provide complementary examples of the same removal task, helping the model learn a consistent residual mapping across different restoration states. The simultaneous improvements in masked and unmasked fidelity show that this training design benefits both overlay removal and background preservation.

We further examine the role of explicit overlay conditioning. In Exp.~2, we replace the overlay conditioning channels with zeros at internal timesteps,
\begin{equation}
  \widehat r=
  \begin{cases}
    v_\theta(z_t,t,[M_s,z_1],c), & t=1000, \\[2pt]
    v_\theta(z_t,t,[M_s,\mathbf{0}],c), & t\in\{250,500,750\},
  \end{cases}
  \label{eq:ocrm_no_internal_overlay}
\end{equation}
where $\mathbf{0}$ has the same shape as $z_1$. The mask channels, text condition, and endpoint conditioning are retained. Compared with this variant, full OCRM improves masked PSNR by $2.57$~dB, reduces masked LPIPS from $0.083$ to $0.074$, and lowers VFID from $0.085$ to $0.076$. These results establish the practical value of retaining the overlay reference during training at internal timesteps.

The role of this reference follows from Eq.~\eqref{eq:ocrm_residual_interpolation}. As $\sigma_t$ changes, the intermediate state contains a varying amount of the original residual, whereas the prediction target remains the complete residual $r=z_1-z_0$. Providing $z_1$ preserves direct access to the original overlay appearance and scene structure throughout this process. The predictor can therefore relate each intermediate state to the same original video when estimating the complete removal residual. This reference also maintains a consistent conditioning structure between internal training samples and endpoint inference. Our design couples residual supervision across different states with a stable visual reference, improving the transfer of training supervision to the deployed endpoint.

The two ablations also reveal complementary performance patterns. Endpoint-only training achieves higher masked PSNR than Exp.~2, while Exp.~2 achieves higher unmasked PSNR and lower VFID. Full OCRM combines stronger regional removal with improved background and perceptual fidelity, reaching $34.38$~dB masked PSNR, $36.59$~dB unmasked PSNR, and $0.076$ VFID. These results demonstrate the complementary roles of supervision at internal timesteps and explicit overlay conditioning in our residual matching design.

\paragraph{Number of Inference Steps.}
We evaluate one to four inference steps on the same 1,000 synthetic videos in Tab.~\ref{tab:step_ablation}. One-step inference achieves the best scores on all three temporal metrics and ties two-step inference on both LPIPS metrics. The largest gains from additional steps are $0.20$~dB in masked PSNR and $0.001$ in VFID, so we use one step by default. This is possible because overlay-to-clean residual matching learns a velocity field that connects the latents of paired overlaid and clean videos. 

\section{More Quantitative Results} 
\label{more_quantitative_results}
Tables~\ref{tab:normal_and_rotated_text}--\ref{tab:curved_text} report results for the ten synthetic subsets of SmartEraser-Bench. 

\section{More Qualitative Results}
\begin{figure*}[!t]
  \centering
  \includegraphics[width=0.92\textwidth]{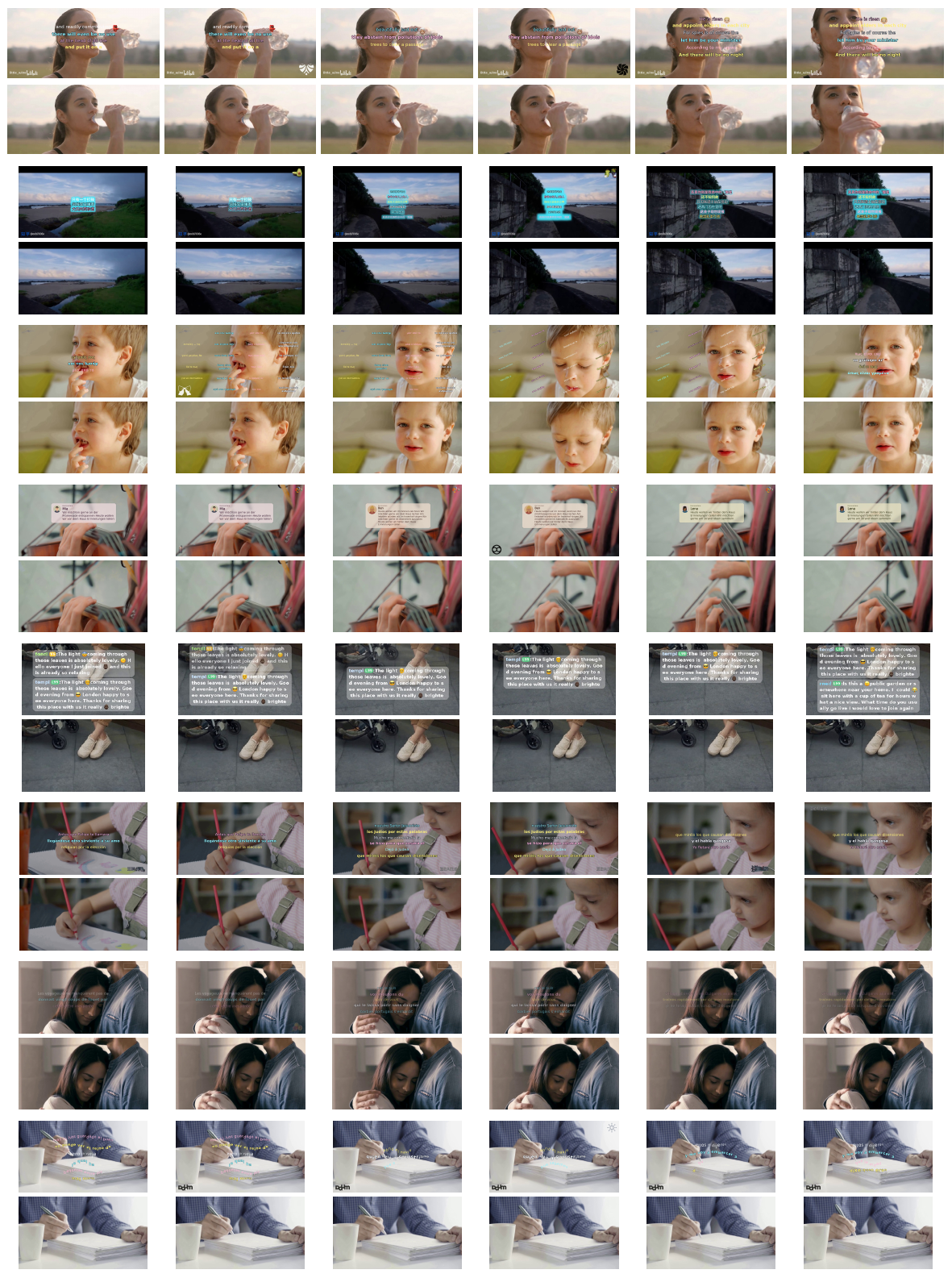}
  \caption{Visualization results on the SmartEraser-Bench synthetic videos. In each pair of rows, the first row shows the original frames, and the second row shows the results of overlay removal by SmartEraser.}
  \label{fig:sys_compact}
\end{figure*}

\begin{figure*}[!t]
  \centering
  \includegraphics[width=0.92\textwidth]{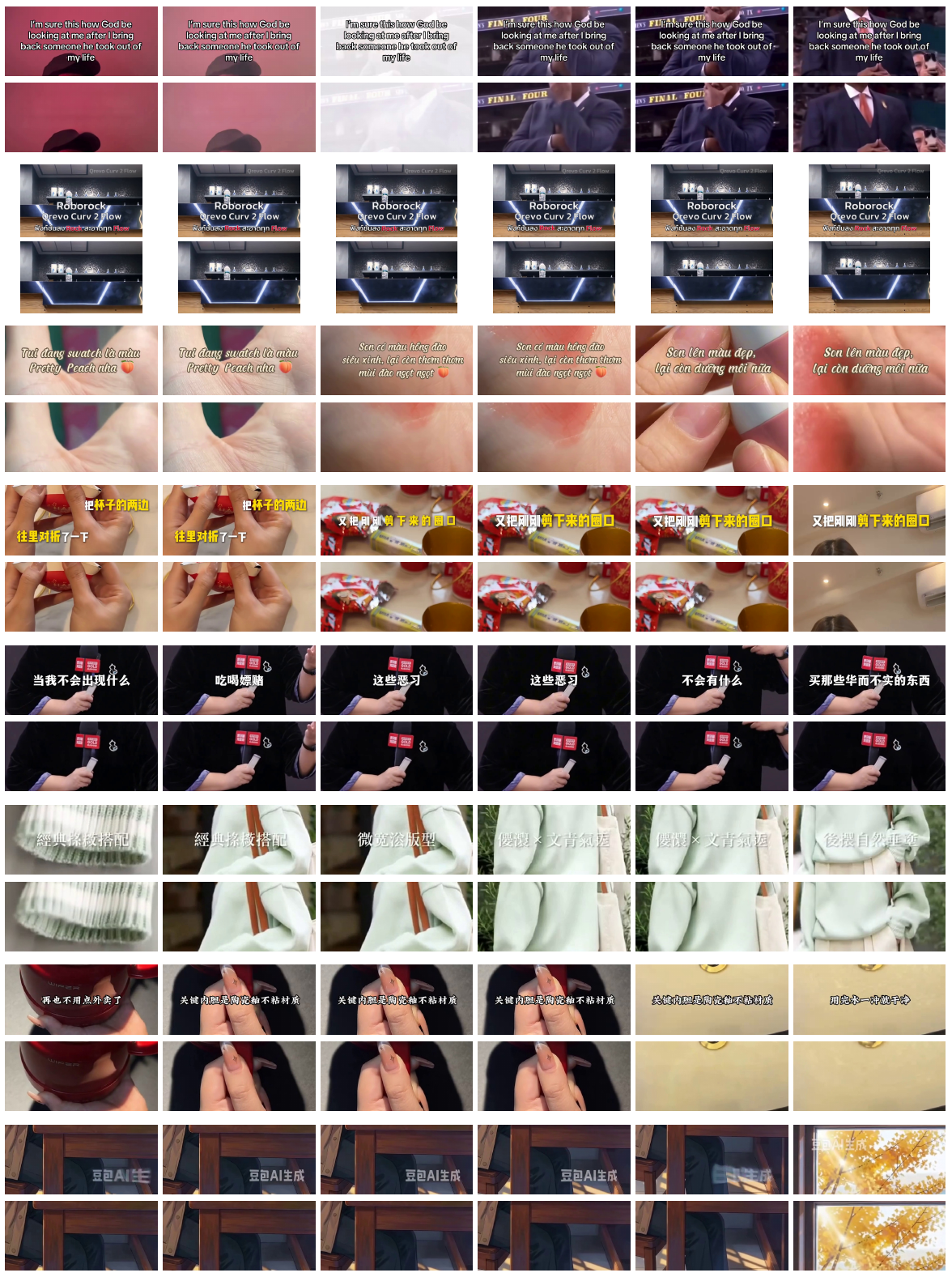}
  \caption{Visualization results on the SmartEraser-Bench real-world videos. In each pair of rows, the first row shows the original frames, and the second row shows the results of overlay removal by SmartEraser.}
  \label{fig:real_compact}
\end{figure*}

As shown in Fig.~\ref{fig:sys_compact} and Fig.~\ref{fig:real_compact}, we show several randomly selected samples of SmartEraser below.

\begin{table*}[!t]
  \centering
  \caption{\textbf{Quantitative evaluation on the \red{Emoji and Sticker} subset of SmartEraser-Bench synthetic videos.}
    $^{\spadesuit}$ denotes fine-tuning on our synthetic dataset. The best and second-best results are in \best{bold} and \second{underlined}, respectively.}
  \label{tab:emoji}
  \resizebox{\textwidth}{!}{%
    \begin{tabular}{llcccccccc}
      \toprule
      & &
      \multicolumn{2}{c}{Overlay Removal} &
      \multicolumn{2}{c}{Background Preservation} &
      \multicolumn{3}{c}{Temporal Consistency} & \\
      \cmidrule(lr){3-4}
      \cmidrule(lr){5-6}
      \cmidrule(lr){7-9}
      Method & Venues
      & \shortstack{Masked\\PSNR $\uparrow$}
      & \shortstack{Masked\\LPIPS $\downarrow$}
      & \shortstack{Unmasked\\PSNR $\uparrow$}
      & \shortstack{Unmasked\\LPIPS $\downarrow$}
      & \shortstack{TC $\downarrow$\\$\times 10^3$}
      & \shortstack{TWE $\downarrow$\\$\times 10^3$}
      & \shortstack{Flow\\Mean $\downarrow$}
      & VFID $\downarrow$ \\
      \midrule
      DiffuEraser \cite{li2025diffueraser} & 2025-arXiv & 31.56 & 0.104 & \second{34.51} & \second{0.057} & \second{5.985} & 0.135 & 0.475 & 0.261 \\
      DiffuEraser$^{\spadesuit}$ \cite{li2025diffueraser} & 2025-arXiv & 31.23 & 0.122 & \second{34.51} & 0.061 & 6.173 & 0.141 & 0.474 & 0.421 \\
      MiniMax-Remover \cite{zi2026minimax} & 2025-NeurIPS & \second{31.92} & \second{0.100} & 33.89 & \second{0.057} & 6.429 & \second{0.127} & \best{0.448} & \second{0.248} \\
      VideoPainter \cite{bian2025videopainter} & 2025-SIGGRAPH & 12.85 & 0.662 & 32.44 & 0.140 & 8.546 & 0.758 & 0.516 & 4.584 \\
      VideoPainter$^{\spadesuit}$ \cite{bian2025videopainter} & 2025-SIGGRAPH & 14.41 & 0.596 & 32.43 & 0.132 & 8.762 & 0.721 & 0.487 & 4.280 \\
      CLEAR \cite{he2026clear} & 2026-ICML & 13.47 & 0.661 & 22.28 & 0.204 & 9.822 & 0.836 & 0.469 & 4.278 \\
      CLEAR$^{\spadesuit}$ \cite{he2026clear} & 2026-ICML & 21.51 & 0.176 & 21.33 & 0.163 & 9.641 & 0.201 & 0.503 & 1.600 \\
      \midrule
      \rowcolor{oursrow}
      SmartEraser (Ours) & 2026 & \best{34.51} & \best{0.071} & \best{36.48} & \best{0.048} & \best{5.949} & \best{0.116} & \second{0.456} & \best{0.166} \\
      \bottomrule
    \end{tabular}
  }
\end{table*}

\begin{table*}[!t]
  \centering
  \caption{\textbf{Quantitative evaluation on the \red{Glowing Text} subset of SmartEraser-Bench synthetic videos.}
    $^{\spadesuit}$ denotes fine-tuning on our synthetic dataset. The best and second-best results are in \best{bold} and \second{underlined}, respectively.}
  \label{tab:glowing_text}
  \resizebox{\textwidth}{!}{%
    \begin{tabular}{llcccccccc}
      \toprule
      & &
      \multicolumn{2}{c}{Overlay Removal} &
      \multicolumn{2}{c}{Background Preservation} &
      \multicolumn{3}{c}{Temporal Consistency} & \\
      \cmidrule(lr){3-4}
      \cmidrule(lr){5-6}
      \cmidrule(lr){7-9}
      Method & Venues
      & \shortstack{Masked\\PSNR $\uparrow$}
      & \shortstack{Masked\\LPIPS $\downarrow$}
      & \shortstack{Unmasked\\PSNR $\uparrow$}
      & \shortstack{Unmasked\\LPIPS $\downarrow$}
      & \shortstack{TC $\downarrow$\\$\times 10^3$}
      & \shortstack{TWE $\downarrow$\\$\times 10^3$}
      & \shortstack{Flow\\Mean $\downarrow$}
      & VFID $\downarrow$ \\
      \midrule
      DiffuEraser \cite{li2025diffueraser} & 2025-arXiv & 17.29 & 0.461 & 25.89 & 0.107 & 7.594 & 0.375 & 0.589 & 2.761 \\
      DiffuEraser$^{\spadesuit}$ \cite{li2025diffueraser} & 2025-arXiv & \second{26.07} & 0.288 & \second{29.79} & \second{0.093} & \second{7.073} & 0.237 & 0.608 & 1.877 \\
      MiniMax-Remover \cite{zi2026minimax} & 2025-NeurIPS & 11.64 & 0.539 & 24.35 & 0.110 & 8.391 & 0.361 & \second{0.579} & 2.783 \\
      VideoPainter \cite{bian2025videopainter} & 2025-SIGGRAPH & 12.14 & 0.740 & 25.46 & 0.161 & 9.271 & 0.879 & \second{0.579} & 5.112 \\
      VideoPainter$^{\spadesuit}$ \cite{bian2025videopainter} & 2025-SIGGRAPH & 12.79 & 0.715 & 25.74 & 0.156 & 9.562 & 0.798 & \best{0.557} & 5.068 \\
      CLEAR \cite{he2026clear} & 2026-ICML & 9.79 & 0.832 & 19.72 & 0.212 & 11.513 & 0.821 & 0.591 & 4.836 \\
      CLEAR$^{\spadesuit}$ \cite{he2026clear} & 2026-ICML & 20.74 & \second{0.207} & 21.02 & 0.165 & 10.292 & \second{0.213} & 0.641 & \second{1.745} \\
      \midrule
      \rowcolor{oursrow}
      SmartEraser (Ours) & 2026 & \best{32.49} & \best{0.104} & \best{36.40} & \best{0.047} & \best{6.387} & \best{0.123} & 0.586 & \best{0.271} \\
      \bottomrule
    \end{tabular}
  }
\end{table*}

\begin{table*}[!t]
  \centering
  \caption{\textbf{Quantitative evaluation on the \red{Full-Screen Text} subset of SmartEraser-Bench synthetic videos.}
    $^{\spadesuit}$ denotes fine-tuning on our synthetic dataset. The best and second-best results are in \best{bold} and \second{underlined}, respectively.}
  \label{tab:full_screen_text}
  \resizebox{\textwidth}{!}{%
    \begin{tabular}{llcccccccc}
      \toprule
      & &
      \multicolumn{2}{c}{Overlay Removal} &
      \multicolumn{2}{c}{Background Preservation} &
      \multicolumn{3}{c}{Temporal Consistency} & \\
      \cmidrule(lr){3-4}
      \cmidrule(lr){5-6}
      \cmidrule(lr){7-9}
      Method & Venues
      & \shortstack{Masked\\PSNR $\uparrow$}
      & \shortstack{Masked\\LPIPS $\downarrow$}
      & \shortstack{Unmasked\\PSNR $\uparrow$}
      & \shortstack{Unmasked\\LPIPS $\downarrow$}
      & \shortstack{TC $\downarrow$\\$\times 10^3$}
      & \shortstack{TWE $\downarrow$\\$\times 10^3$}
      & \shortstack{Flow\\Mean $\downarrow$}
      & VFID $\downarrow$ \\
      \midrule
      DiffuEraser \cite{li2025diffueraser} & 2025-arXiv & 32.03 & 0.091 & 34.82 & 0.058 & \second{5.769} & 0.113 & 0.413 & 0.320 \\
      DiffuEraser$^{\spadesuit}$ \cite{li2025diffueraser} & 2025-arXiv & 31.70 & 0.108 & \second{34.85} & 0.066 & 6.130 & 0.126 & 0.414 & 0.605 \\
      MiniMax-Remover \cite{zi2026minimax} & 2025-NeurIPS & \second{32.46} & \second{0.083} & 34.48 & \second{0.054} & 5.917 & \second{0.109} & \second{0.384} & \second{0.316} \\
      VideoPainter \cite{bian2025videopainter} & 2025-SIGGRAPH & 12.72 & 0.709 & 32.11 & 0.247 & 9.558 & 1.231 & 0.475 & 7.751 \\
      VideoPainter$^{\spadesuit}$ \cite{bian2025videopainter} & 2025-SIGGRAPH & 14.99 & 0.555 & 32.74 & 0.208 & 9.594 & 1.129 & 0.458 & 6.821 \\
      CLEAR \cite{he2026clear} & 2026-ICML & 12.22 & 0.829 & 23.10 & 0.321 & 9.518 & 1.314 & \best{0.380} & 7.362 \\
      CLEAR$^{\spadesuit}$ \cite{he2026clear} & 2026-ICML & 21.32 & 0.167 & 21.40 & 0.157 & 8.451 & 0.171 & 0.438 & 1.722 \\
      \midrule
      \rowcolor{oursrow}
      SmartEraser (Ours) & 2026 & \best{35.64} & \best{0.057} & \best{36.93} & \best{0.042} & \best{5.434} & \best{0.094} & 0.393 & \best{0.183} \\
      \bottomrule
    \end{tabular}
  }
\end{table*}

\begin{table*}[!t]
  \centering
  \caption{\textbf{Quantitative evaluation on the \red{Comment Card} subset of SmartEraser-Bench synthetic videos.}
    $^{\spadesuit}$ denotes fine-tuning on our synthetic dataset. The best and second-best results are in \best{bold} and \second{underlined}, respectively.}
  \label{tab:comment_card}
  \resizebox{\textwidth}{!}{%
    \begin{tabular}{llcccccccc}
      \toprule
      & &
      \multicolumn{2}{c}{Overlay Removal} &
      \multicolumn{2}{c}{Background Preservation} &
      \multicolumn{3}{c}{Temporal Consistency} & \\
      \cmidrule(lr){3-4}
      \cmidrule(lr){5-6}
      \cmidrule(lr){7-9}
      Method & Venues
      & \shortstack{Masked\\PSNR $\uparrow$}
      & \shortstack{Masked\\LPIPS $\downarrow$}
      & \shortstack{Unmasked\\PSNR $\uparrow$}
      & \shortstack{Unmasked\\LPIPS $\downarrow$}
      & \shortstack{TC $\downarrow$\\$\times 10^3$}
      & \shortstack{TWE $\downarrow$\\$\times 10^3$}
      & \shortstack{Flow\\Mean $\downarrow$}
      & VFID $\downarrow$ \\
      \midrule
      DiffuEraser \cite{li2025diffueraser} & 2025-arXiv & 28.41 & 0.126 & 34.56 & \second{0.052} & \second{6.615} & 0.167 & 0.509 & \second{0.365} \\
      DiffuEraser$^{\spadesuit}$ \cite{li2025diffueraser} & 2025-arXiv & 28.07 & 0.135 & \second{34.57} & 0.055 & 6.849 & 0.175 & 0.508 & 0.498 \\
      MiniMax-Remover \cite{zi2026minimax} & 2025-NeurIPS & \second{29.37} & \second{0.119} & 33.69 & \second{0.052} & 6.965 & \second{0.154} & \best{0.486} & 0.366 \\
      VideoPainter \cite{bian2025videopainter} & 2025-SIGGRAPH & 14.84 & 0.515 & 32.57 & 0.119 & 9.628 & 0.752 & 0.522 & 2.857 \\
      VideoPainter$^{\spadesuit}$ \cite{bian2025videopainter} & 2025-SIGGRAPH & 16.08 & 0.497 & 32.67 & 0.113 & 10.374 & 0.771 & 0.532 & 2.907 \\
      CLEAR \cite{he2026clear} & 2026-ICML & 8.48 & 0.802 & 23.25 & 0.168 & 11.451 & 1.283 & 0.519 & 4.358 \\
      CLEAR$^{\spadesuit}$ \cite{he2026clear} & 2026-ICML & 21.63 & 0.171 & 21.55 & 0.148 & 10.203 & 0.222 & 0.542 & 1.479 \\
      \midrule
      \rowcolor{oursrow}
      SmartEraser (Ours) & 2026 & \best{33.43} & \best{0.083} & \best{36.19} & \best{0.045} & \best{6.332} & \best{0.136} & \second{0.492} & \best{0.219} \\
      \bottomrule
    \end{tabular}
  }
\end{table*}

\begin{table*}[!t]
  \centering
  \caption{\textbf{Quantitative evaluation on the \red{Transition Blur} subset of SmartEraser-Bench synthetic videos.}
    $^{\spadesuit}$ denotes fine-tuning on our synthetic dataset. The best and second-best results are in \best{bold} and \second{underlined}, respectively.}
  \label{tab:transition_blur}
  \resizebox{\textwidth}{!}{%
    \begin{tabular}{llcccccccc}
      \toprule
      & &
      \multicolumn{2}{c}{Overlay Removal} &
      \multicolumn{2}{c}{Background Preservation} &
      \multicolumn{3}{c}{Temporal Consistency} & \\
      \cmidrule(lr){3-4}
      \cmidrule(lr){5-6}
      \cmidrule(lr){7-9}
      Method & Venues
      & \shortstack{Masked\\PSNR $\uparrow$}
      & \shortstack{Masked\\LPIPS $\downarrow$}
      & \shortstack{Unmasked\\PSNR $\uparrow$}
      & \shortstack{Unmasked\\LPIPS $\downarrow$}
      & \shortstack{TC $\downarrow$\\$\times 10^3$}
      & \shortstack{TWE $\downarrow$\\$\times 10^3$}
      & \shortstack{Flow\\Mean $\downarrow$}
      & VFID $\downarrow$ \\
      \midrule
      DiffuEraser \cite{li2025diffueraser} & 2025-arXiv & \second{26.56} & \second{0.137} & \second{34.47} & \second{0.056} & \second{8.374} & 2.938 & 1.335 & \second{0.288} \\
      DiffuEraser$^{\spadesuit}$ \cite{li2025diffueraser} & 2025-arXiv & 26.14 & 0.150 & 34.46 & 0.059 & 8.605 & 2.993 & 1.322 & 0.351 \\
      MiniMax-Remover \cite{zi2026minimax} & 2025-NeurIPS & 24.93 & 0.145 & 32.76 & 0.061 & 10.834 & \second{2.714} & \best{1.230} & 0.319 \\
      VideoPainter \cite{bian2025videopainter} & 2025-SIGGRAPH & 12.48 & 0.634 & 29.75 & 0.136 & 19.275 & 3.780 & 1.428 & 3.313 \\
      VideoPainter$^{\spadesuit}$ \cite{bian2025videopainter} & 2025-SIGGRAPH & 13.51 & 0.576 & 29.29 & 0.133 & 19.225 & 3.470 & 1.267 & 3.107 \\
      CLEAR \cite{he2026clear} & 2026-ICML & 14.69 & 0.602 & 22.61 & 0.184 & 20.225 & 4.286 & 1.694 & 2.710 \\
      CLEAR$^{\spadesuit}$ \cite{he2026clear} & 2026-ICML & 21.27 & 0.193 & 22.01 & 0.162 & 16.918 & \best{2.540} & 1.375 & 1.097 \\
      \midrule
      \rowcolor{oursrow}
      SmartEraser (Ours) & 2026 & \best{32.06} & \best{0.086} & \best{35.86} & \best{0.046} & \best{7.870} & 2.895 & \second{1.263} & \best{0.193} \\
      \bottomrule
    \end{tabular}
  }
\end{table*}

\begin{table*}[!t]
  \centering
  \caption{\textbf{Quantitative evaluation on the \red{Live Chat} subset of SmartEraser-Bench synthetic videos.}
    $^{\spadesuit}$ denotes fine-tuning on our synthetic dataset. The best and second-best results are in \best{bold} and \second{underlined}, respectively.}
  \label{tab:live_comments}
  \resizebox{\textwidth}{!}{%
    \begin{tabular}{llcccccccc}
      \toprule
      & &
      \multicolumn{2}{c}{Overlay Removal} &
      \multicolumn{2}{c}{Background Preservation} &
      \multicolumn{3}{c}{Temporal Consistency} & \\
      \cmidrule(lr){3-4}
      \cmidrule(lr){5-6}
      \cmidrule(lr){7-9}
      Method & Venues
      & \shortstack{Masked\\PSNR $\uparrow$}
      & \shortstack{Masked\\LPIPS $\downarrow$}
      & \shortstack{Unmasked\\PSNR $\uparrow$}
      & \shortstack{Unmasked\\LPIPS $\downarrow$}
      & \shortstack{TC $\downarrow$\\$\times 10^3$}
      & \shortstack{TWE $\downarrow$\\$\times 10^3$}
      & \shortstack{Flow\\Mean $\downarrow$}
      & VFID $\downarrow$ \\
      \midrule
      DiffuEraser \cite{li2025diffueraser} & 2025-arXiv & 26.19 & 0.195 & \second{34.01} & \second{0.055} & \second{6.447} & 0.196 & 0.455 & \second{0.791} \\
      DiffuEraser$^{\spadesuit}$ \cite{li2025diffueraser} & 2025-arXiv & 25.70 & 0.203 & 33.95 & 0.058 & 6.745 & 0.205 & 0.453 & 0.905 \\
      MiniMax-Remover \cite{zi2026minimax} & 2025-NeurIPS & \second{27.27} & \second{0.183} & 33.64 & \second{0.055} & 6.627 & \second{0.171} & \best{0.423} & 0.834 \\
      VideoPainter \cite{bian2025videopainter} & 2025-SIGGRAPH & 15.59 & 0.520 & 32.02 & 0.132 & 10.109 & 0.711 & 0.634 & 5.102 \\
      VideoPainter$^{\spadesuit}$ \cite{bian2025videopainter} & 2025-SIGGRAPH & 15.92 & 0.558 & 31.40 & 0.131 & 11.139 & 0.767 & 0.525 & 5.110 \\
      CLEAR \cite{he2026clear} & 2026-ICML & 10.90 & 0.762 & 22.14 & 0.198 & 16.105 & 1.251 & 0.519 & 6.269 \\
      CLEAR$^{\spadesuit}$ \cite{he2026clear} & 2026-ICML & 20.99 & 0.194 & 20.57 & 0.164 & 9.887 & 0.250 & 0.485 & 1.914 \\
      \midrule
      \rowcolor{oursrow}
      SmartEraser (Ours) & 2026 & \best{35.24} & \best{0.081} & \best{36.36} & \best{0.044} & \best{5.897} & \best{0.156} & \second{0.444} & \best{0.231} \\
      \bottomrule
    \end{tabular}
  }
\end{table*}

\begin{table*}[!t]
  \centering
  \caption{\textbf{Quantitative evaluation on the \red{Dynamic Watermark} subset of SmartEraser-Bench synthetic videos.}
    $^{\spadesuit}$ denotes fine-tuning on our synthetic dataset. The best and second-best results are in \best{bold} and \second{underlined}, respectively.}
  \label{tab:dynamic_watermark}
  \resizebox{\textwidth}{!}{%
    \begin{tabular}{llcccccccc}
      \toprule
      & &
      \multicolumn{2}{c}{Overlay Removal} &
      \multicolumn{2}{c}{Background Preservation} &
      \multicolumn{3}{c}{Temporal Consistency} & \\
      \cmidrule(lr){3-4}
      \cmidrule(lr){5-6}
      \cmidrule(lr){7-9}
      Method & Venues
      & \shortstack{Masked\\PSNR $\uparrow$}
      & \shortstack{Masked\\LPIPS $\downarrow$}
      & \shortstack{Unmasked\\PSNR $\uparrow$}
      & \shortstack{Unmasked\\LPIPS $\downarrow$}
      & \shortstack{TC $\downarrow$\\$\times 10^3$}
      & \shortstack{TWE $\downarrow$\\$\times 10^3$}
      & \shortstack{Flow\\Mean $\downarrow$}
      & VFID $\downarrow$ \\
      \midrule
      DiffuEraser \cite{li2025diffueraser} & 2025-arXiv & \second{31.46} & 0.112 & \second{34.75} & 0.057 & \best{6.243} & 0.169 & 0.488 & \second{0.295} \\
      DiffuEraser$^{\spadesuit}$ \cite{li2025diffueraser} & 2025-arXiv & 31.04 & 0.129 & \second{34.75} & 0.060 & 6.430 & 0.176 & 0.488 & 0.432 \\
      MiniMax-Remover \cite{zi2026minimax} & 2025-NeurIPS & 31.30 & \second{0.111} & 34.15 & \second{0.056} & 6.688 & \second{0.160} & \best{0.463} & 0.312 \\
      VideoPainter \cite{bian2025videopainter} & 2025-SIGGRAPH & 13.10 & 0.617 & 32.21 & 0.131 & 8.970 & 0.755 & 0.504 & 5.304 \\
      VideoPainter$^{\spadesuit}$ \cite{bian2025videopainter} & 2025-SIGGRAPH & 14.32 & 0.573 & 31.83 & 0.127 & 9.223 & 0.713 & 0.474 & 4.988 \\
      CLEAR \cite{he2026clear} & 2026-ICML & 15.53 & 0.578 & 22.10 & 0.193 & 10.197 & 0.874 & 0.489 & 4.217 \\
      CLEAR$^{\spadesuit}$ \cite{he2026clear} & 2026-ICML & 21.60 & 0.188 & 21.61 & 0.169 & 10.026 & 0.240 & 0.526 & 2.082 \\
      \midrule
      \rowcolor{oursrow}
      SmartEraser (Ours) & 2026 & \best{34.98} & \best{0.073} & \best{36.98} & \best{0.048} & \second{6.246} & \best{0.150} & \second{0.470} & \best{0.189} \\
      \bottomrule
    \end{tabular}
  }
\end{table*}

\begin{table*}[!t]
  \centering
  \caption{\textbf{Quantitative evaluation on the \red{Transparent Text} subset of SmartEraser-Bench synthetic videos.}
    $^{\spadesuit}$ denotes fine-tuning on our synthetic dataset. The best and second-best results are in \best{bold} and \second{underlined}, respectively.}
  \label{tab:transparent_text}
  \resizebox{\textwidth}{!}{%
    \begin{tabular}{llcccccccc}
      \toprule
      & &
      \multicolumn{2}{c}{Overlay Removal} &
      \multicolumn{2}{c}{Background Preservation} &
      \multicolumn{3}{c}{Temporal Consistency} & \\
      \cmidrule(lr){3-4}
      \cmidrule(lr){5-6}
      \cmidrule(lr){7-9}
      Method & Venues
      & \shortstack{Masked\\PSNR $\uparrow$}
      & \shortstack{Masked\\LPIPS $\downarrow$}
      & \shortstack{Unmasked\\PSNR $\uparrow$}
      & \shortstack{Unmasked\\LPIPS $\downarrow$}
      & \shortstack{TC $\downarrow$\\$\times 10^3$}
      & \shortstack{TWE $\downarrow$\\$\times 10^3$}
      & \shortstack{Flow\\Mean $\downarrow$}
      & VFID $\downarrow$ \\
      \midrule
      DiffuEraser \cite{li2025diffueraser} & 2025-arXiv & \second{29.80} & \second{0.122} & \second{34.58} & \second{0.057} & \second{6.226} & 0.177 & 0.572 & \second{0.349} \\
      DiffuEraser$^{\spadesuit}$ \cite{li2025diffueraser} & 2025-arXiv & 29.49 & 0.138 & \second{34.58} & 0.060 & 6.388 & 0.182 & 0.571 & 0.485 \\
      MiniMax-Remover \cite{zi2026minimax} & 2025-NeurIPS & 29.49 & 0.127 & 33.87 & 0.058 & 6.810 & \second{0.172} & \second{0.549} & 0.428 \\
      VideoPainter \cite{bian2025videopainter} & 2025-SIGGRAPH & 13.89 & 0.610 & 32.41 & 0.135 & 8.522 & 0.629 & 0.574 & 4.804 \\
      VideoPainter$^{\spadesuit}$ \cite{bian2025videopainter} & 2025-SIGGRAPH & 15.05 & 0.568 & 32.24 & 0.130 & 8.790 & 0.607 & \best{0.547} & 4.723 \\
      CLEAR \cite{he2026clear} & 2026-ICML & 16.04 & 0.645 & 21.61 & 0.193 & 10.484 & 0.558 & 0.588 & 4.454 \\
      CLEAR$^{\spadesuit}$ \cite{he2026clear} & 2026-ICML & 21.34 & 0.172 & 21.55 & 0.157 & 10.329 & 0.270 & 0.606 & 1.867 \\
      \midrule
      \rowcolor{oursrow}
      SmartEraser (Ours) & 2026 & \best{34.45} & \best{0.070} & \best{36.51} & \best{0.047} & \best{6.160} & \best{0.156} & 0.559 & \best{0.163} \\
      \bottomrule
    \end{tabular}
  }
\end{table*}

\begin{table*}[!t]
  \centering
  \caption{\textbf{Quantitative evaluation on the \red{Curved Text} subset of SmartEraser-Bench synthetic videos.}
    $^{\spadesuit}$ denotes fine-tuning on our synthetic dataset. The best and second-best results are in \best{bold} and \second{underlined}, respectively.}
  \label{tab:curved_text}
  \resizebox{\textwidth}{!}{%
    \begin{tabular}{llcccccccc}
      \toprule
      & &
      \multicolumn{2}{c}{Overlay Removal} &
      \multicolumn{2}{c}{Background Preservation} &
      \multicolumn{3}{c}{Temporal Consistency} & \\
      \cmidrule(lr){3-4}
      \cmidrule(lr){5-6}
      \cmidrule(lr){7-9}
      Method & Venues
      & \shortstack{Masked\\PSNR $\uparrow$}
      & \shortstack{Masked\\LPIPS $\downarrow$}
      & \shortstack{Unmasked\\PSNR $\uparrow$}
      & \shortstack{Unmasked\\LPIPS $\downarrow$}
      & \shortstack{TC $\downarrow$\\$\times 10^3$}
      & \shortstack{TWE $\downarrow$\\$\times 10^3$}
      & \shortstack{Flow\\Mean $\downarrow$}
      & VFID $\downarrow$ \\
      \midrule
      DiffuEraser \cite{li2025diffueraser} & 2025-arXiv & 31.35 & 0.097 & \second{35.03} & 0.052 & \second{5.640} & 0.147 & 0.480 & 0.324 \\
      DiffuEraser$^{\spadesuit}$ \cite{li2025diffueraser} & 2025-arXiv & 30.97 & 0.113 & 35.01 & 0.055 & 5.866 & 0.154 & 0.479 & 0.551 \\
      MiniMax-Remover \cite{zi2026minimax} & 2025-NeurIPS & \second{31.89} & \second{0.088} & 34.32 & \second{0.050} & 5.991 & \second{0.140} & \best{0.455} & \second{0.292} \\
      VideoPainter \cite{bian2025videopainter} & 2025-SIGGRAPH & 13.00 & 0.653 & 32.74 & 0.132 & 8.585 & 0.885 & 0.519 & 4.913 \\
      VideoPainter$^{\spadesuit}$ \cite{bian2025videopainter} & 2025-SIGGRAPH & 15.36 & 0.558 & 32.75 & 0.123 & 8.740 & 0.803 & 0.481 & 4.528 \\
      CLEAR \cite{he2026clear} & 2026-ICML & 13.51 & 0.710 & 22.45 & 0.198 & 9.413 & 0.975 & \second{0.464} & 4.732 \\
      CLEAR$^{\spadesuit}$ \cite{he2026clear} & 2026-ICML & 20.80 & 0.187 & 20.97 & 0.174 & 9.349 & 0.227 & 0.503 & 1.985 \\
      \midrule
      \rowcolor{oursrow}
      SmartEraser (Ours) & 2026 & \best{35.83} & \best{0.056} & \best{37.15} & \best{0.040} & \best{5.458} & \best{0.128} & 0.465 & \best{0.166} \\
      \bottomrule
    \end{tabular}
  }
\end{table*}

\end{document}